\documentclass[twoside]{article}

\usepackage[accepted]{aistats2020}

\usepackage[round]{natbib}

\usepackage{hyperref}       
\usepackage{url}            
\usepackage{booktabs}       
\usepackage{amsfonts}       
\usepackage{nicefrac}       
\usepackage{microtype}      
\usepackage{xspace}
\usepackage{graphicx}
\usepackage{multirow}
\usepackage{wrapfig,lipsum,booktabs}
\usepackage{sidecap}
\usepackage{lmodern}

\usepackage{array}
\usepackage[dvipsnames]{xcolor}
\newcommand*{\rowstyle}[1]{
	\gdef\@rowstyle{#1}%
	\@rowstyle\ignorespaces%
}
\newcolumntype{=}{
	>{\gdef\@rowstyle{}}%
}
\newcolumntype{+}{
	>{\@rowstyle}%
}

\usepackage{caption}
\usepackage{subcaption} 
\usepackage[ruled, vlined]{algorithm2e}
\usepackage{paralist}
\usepackage{amsmath,amsfonts}
\usepackage{amssymb,amsopn}
\usepackage{bm} 

\usepackage{amssymb}
\usepackage{amsmath,amsfonts}
\usepackage{amsthm,amsopn}

\usepackage{bm} 

\newcommand{\field}[1]{\mathbb{#1}}
\newcommand{\R}{\field{R}} 

\newcommand{\ProbOpr}[1]{\mathbb{#1}}

\newcommand{\expect}[2]{%
\ifthenelse{\equal{#2}{}}{\ProbOpr{E}_{#1}}
{\ifthenelse{\equal{#1}{}}{\ProbOpr{E}\left[#2\right]}{\ProbOpr{E}_{#1}\left[#2\right]}}} 
\newcommand{\var}[2]{%
\ifthenelse{\equal{#2}{}}{\ProbOpr{VAR}_{#1}}
{\ifthenelse{\equal{#1}{}}{\ProbOpr{VAR}\left[#2\right]}{\ProbOpr{VAR}_{#1}\left[#2\right]}}} 

\DeclareMathOperator{\argmax}{arg\,max}
\DeclareMathOperator{\argmin}{arg\,min}

\newcommand{\val}{\text{val}}
\newcommand{\tr}{\text{tr}}
\newcommand{\ML}{\text{meta}}
\newcommand{\sH}{\mathcal{H}}

\newcommand{\sY}{\mathcal{Y}}
\newcommand{\sX}{\mathcal{X}}

\newcommand{\sD}{\mathcal{D}}
\newcommand{\sG}{\mathcal{G}}
\newcommand{\sI}{\mathcal{I}}
\newcommand{\sO}{\mathcal{O}}

\newcommand{\eat}[1]{}
\renewcommand{\paragraph}[1]{\vspace{0.0ex}\noindent\textbf{#1}}
\newcommand{\ie}{i.e.\xspace}
\newcommand{\eg}{e.g.\xspace}

\begin{document}
	
\runningauthor{Wei-Lun Chao, Han-Jia Ye, De-Chuan Zhan, Mark Campbell, Kilian Q. Weinberger}
	
\twocolumn[
\aistatstitle{Revisiting Meta-Learning as Supervised Learning}
\aistatsauthor{Wei-Lun Chao$^*$ \And Han-Jia Ye$^*$ \And De-Chuan Zhan}
\aistatsaddress{The Ohio State University, USA \And Nanjing University, China \And Nanjing University, China} 
\aistatsauthor{Mark Campbell \And Kilian Q. Weinberger}
\aistatsaddress{Cornell University, USA \And  Cornell University, USA} 
]
	
\begin{abstract}
Recent years have witnessed an abundance of new publications and approaches on meta-learning. This community-wide enthusiasm has sparked great insights but has also created a plethora of seemingly different frameworks, which can be hard to compare and evaluate.
In this paper, we aim to provide a principled, unifying framework by \emph{revisiting and strengthening the connection between meta-learning and traditional supervised learning.}
By treating pairs of task-specific data sets and target models as (feature, label) samples, we can reduce many meta-learning algorithms to instances of supervised learning.
This view not only unifies meta-learning into an intuitive and practical framework but also allows us to transfer insights from supervised learning directly to improve meta-learning.
For example, we obtain a better understanding of generalization properties, and we can readily transfer well-understood techniques, such as model ensemble, pre-training, joint training, data augmentation, and even nearest neighbor based methods.
We provide an intuitive analogy of these methods in the context of meta-learning and show that they give rise to significant improvements in model performance on few-shot learning. 
\end{abstract}
\section{Introduction}
Meta-learning, or learning to learn, is the  sub-field of machine learning occupied with the search for the best learning strategy as the number of tasks and learning experiences increases~\citep{vilalta2002perspective} and has drawn significant attention recently~\citep{finn2017model,andrychowicz2016learning,vinyals2016matching}. 
Meta-learning has been developed in various areas to advanced algorithm design, including few-shot learning~\citep{ravi2017optimization,snell2017prototypical,wang2016learning}, optimization~\citep{li2017learning,wichrowska2017learned}, active learning~\citep{bachman2017learning}, transfer learning~\citep{balaji2018metareg,ying2018transfer}, unsupervised learning~\citep{metz2018learning, edwards2017towards}, etc. Specifically, meta-learning has demonstrated the capability to generalize learned knowledge to novel tasks, which greatly reduces the need for training data and time to optimize. 

\begin{figure}
	\centering
	\includegraphics[width=0.95\linewidth]{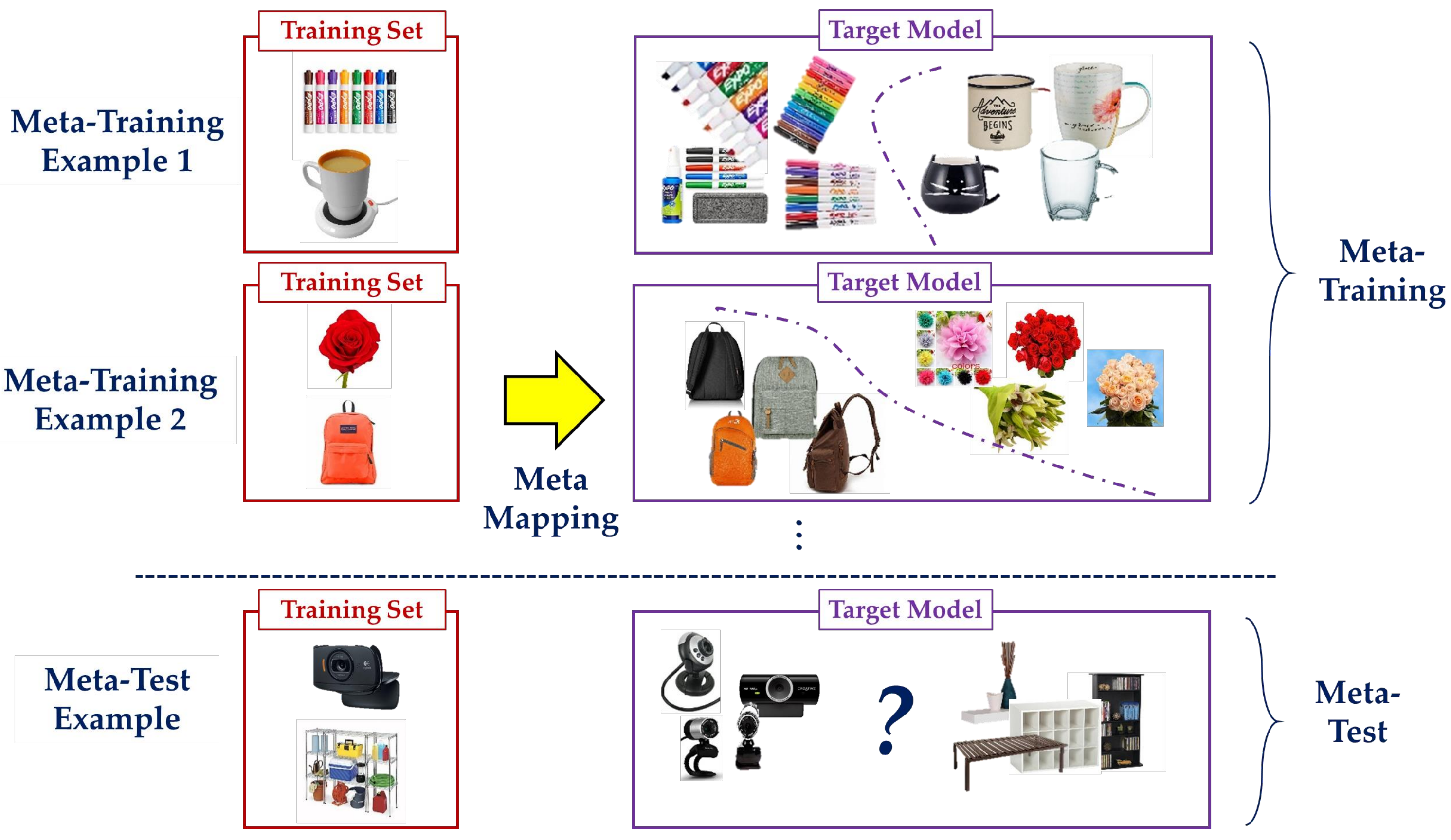}
	\vspace{-10pt}
	\caption{\small \textbf{Meta-learning as supervised learning.} By treating a pair of a training set (left) and the target model (right) as a meta labeled example, meta-learning learns a shared meta mapping (from left to right) that generalizes from meta-training examples to novel meta-test examples. Take one-shot learning for instance. A training set contains one image per class, while a target model is a classifier trained with many more images from those classes.}
	\label{fig:concept}
	\vspace{-10pt}
\end{figure}

Given its wide applicability and diverse approaches, there is an increasing need for a principled, unifying meta-learning framework to facilitate future studies and development. \citet{finn2018learning,metz2018learning} have made a notable step to provide a broad introduction and compare representative algorithms. In this paper, we aim to push the direction forward by providing a unifying view of meta-learning through \emph{revisiting and strengthening the connection to supervised learning}~\citep{thrun1998learning,baxter2000model,maurer2005algorithmic,wang2018learning2,larochelle2018few,finn2018learning}.

Generally, meta-learning can be viewed as learning a mapping $g$ that given a training set $D_\tr$ returns a model $h$. The model $h$ is then applied to data sampled from the same distribution from which $D_\tr$ is sampled.
For instance, in one-shot learning $D_\tr$ corresponds to a set of $C$ labeled data samples, and $h$ corresponds to a $C$-way classifier.
Learning the mapping can thus be conducted by collecting and training with $({D_\tr},{D_\val})$ pairs, in which the validation set $D_\val$ evaluates $h=g(D_\tr)$ and provides ``supervised'' signal to optimize $g$, essentially connecting meta-learning with supervised learning\footnote{Suppose $(x, y)\in D_\val$, where $x$ is the feature and $y$ is the label, \citet{finn2018learning} equivalently defines $h(x)=g(D_\tr, x)$ and uses $y$ to provide the supervised signal.}. This connection has multiple advantages: it allows transferring ideas and experiences of supervised learning to meta-learning. For example, \cite{baxter2000model,maurer2005algorithmic} extended the generalization error bounds from supervised learning to meta-learning; ~\cite{vinyals2016matching} designed the meta-training procedure so that meta-test and meta-train conditions match.

In this paper, we aim to generalize, expand, and strengthen the connection between meta-learning and supervised learning so that on one hand, it can hold across a wider range of meta-learning approaches and applications; on the other hand, a variety of understandings and techniques of supervised learning can be easily extended to meta-learning. 

Traditional supervised learning learns a model that given a feature (vector) returns a label, and we train the model on a set of labeled examples in the form of (feature, target label) pairs. Inspired by this fact and \citep{wang2016learning,garg2018supervising}, we provide a unifying view of meta-learning by drawing a closer analogy. We \emph{pretend} that the target model $h^*$ of a training set $D_\tr$ can be directly obtained in collecting meta-training data. One can view the target model $h^*$ as the model that we hope to learn from $D_\tr$, similar to the annotated label that we hope to predict given a feature. In this notion, we can train a meta model $g$ on a set of $({D_\tr},h^*)$ pairs, which we call the meta labeled examples (or tasks\footnote{There are various definitions of the term ``task.'' For example, in \citep{pan2010survey, finn2018learning}, a task contains a data distribution where the training set $D_\tr$ is sampled, and a task can be sampled from a meta-distribution of tasks. Our definition is equivalent to combining the two sampling steps and can simplify the notations in this paper.}), and evaluate it on a test task $D_\tr^\text{test}$. See \autoref{fig:concept} for an illustration. 

This unifying view
allows us to apply meta-learning to various areas while general understandings of supervised learning still hold. For example, in meta-learning for domain generalization~\citep{balaji2018metareg} where $h=g(D_\tr)$ is applied to a data distribution different from where $D_\tr$ is sampled, the theoretical analysis in~\citep{baxter2000model,maurer2005algorithmic} is no longer applicable. Nevertheless, 
well understood pitfalls of supervised learning such as overfitting due to insufficient (meta-) training examples and distribution drift can still be used to explain why the learned meta model $g$ may not generalize to novel tasks. We discuss more in \autoref{S_general} and empirically verify this in \autoref{S_DGMD}.

This unifying view also identifies the essential components to design a meta-learning algorithm, providing a principled way to apply it. While $h^*$ may not be directly provided in practice, it indicates what kind of data to collect. For example, in one-shot learning one may expect $h^*$ to be a classifier learned with  ample labeled data~\citep{wang2016learning,wang2017learning}. To this end, one can first collect ample labeled data to learn $h^*$, followed by learning a meta-learning model $g$ to predict $h^*$ from $D_\tr$. Alternatively, one can collect a validation set $D_\val$ as a proxy to estimate how $h^*$ will perform. See \autoref{S_appl} for more discussions.

Finally, with this unifying view, we broaden the scope of algorithm design for meta-learning by extending concepts from supervised learning.
We conduct extensive experiments on few-shot learning, a representative area where meta-learning is applied. We empirically show that well-known supervised learning techniques such as data augmentation, bagging~\citep{breiman1996bagging}, joint training~\citep{argyriou2007multi}, pre-training~\citep{yosinski2014transferable}, and non-parametric approaches~\citep{zhang2006svm,weinberger2009distance} can be applied at the task level to significantly facilitate meta-learning.

\section{A Unifying View of Meta-Learning}

\label{S_approach}

\subsection{Background: supervised learning}
\label{S_approach_back}

In supervised learning we collect a training set $D_{\tr}=\{(x_n\in\sX ,y_n\in\sY)\}_{n=1}^N$, composed of $N$ i.i.d. samples from an unknown distribution $\sD$ on $\sX\times\sY$. We call $x$ an input, $y$ an output (or label), and $(x, y)$ a labeled example.  We call $h:\sX\mapsto\sY$ a model, which outputs a label for an input. For instance, in image classification, $x$ is an image, $y$ is a class name (e.g., ``dog''), $\sD$ is the distribution of real images, and $h$ is an image classifier.

Supervised learning searches for a model $h$ given $D_{\tr}$, so that $h$ will work well on $(x,y)$ sampled from $\sD$. That is, $h$ should have a small generalization error $L_\sD(h)$ according to a loss $l: \sY \times\sY \mapsto \R$
\begin{align}
\vspace{-5pt}
L_\sD(h) = E_{(x,y)\sim\sD}[l(h(x), y)].  \label{e_test_error}
\end{align}
To this end, we construct a hypothesis set $\sH=\{h\}$ of candidate models and design an algorithm $A_\sH$ to search $\hat{h}$ from $\sH$ by learning from $D_{\tr}$.
We denote $\hat{h} = A_\sH(D_{\tr})$. One example of $\sH$ is a neural network with a fixed architecture but undetermined weights.

A popular framework to design supervised learning algorithms is empirical risk minimization (ERM)
\vspace{-5pt}
\begin{align}
\hat{h} = A_\sH(D_{\tr}) & = \argmin_{h\in\mathcal{H}} \cfrac{1}{N}\sum_{n=1}^N l(h(x_n), y_n) \nonumber \\ & = \argmin_{h\in\mathcal{H}} L_S(h) \label{e_RLM},
\end{align}
where $L_S(h)$ is the training loss. A particular $A_\sH$ is characterized by how it performs ERM (e.g., optimization methods). In practice, we usually design a set of $(\sH$, $A_\sH)$ pairs, denoted as $G$, and select the best pair using a held-out validation set $D_{\val}=\{(x_m,y_m)\}_{m=1}^M$ sampled i.i.d. from $\sD$ \vspace{-5pt}
\begin{align}
(\sH, A_{\sH})^* & = \argmin_{(\sH, A_\sH)\in G} \cfrac{1}{M}\sum_{m=1}^M l(\hat{h}(x_m), y_m) \nonumber\\
& = \argmin_{(\sH, A_\sH)\in G} L_V(\hat{h}), \label{e_val}
\end{align}
where $\hat{h} = A_{\sH}(D_{\tr})$. This is called model selection and $L_V$ is the validation error.

\subsection{Meta-learning as supervised learning}
\label{S_meta_idea}
We provide a framework of meta-learning by drawing analogy to supervised learning. We use ``meta (labeled) example'' and ``task'' interchangeably. 
To prevent confusion, we call the model in supervised learning a ``base'' model when needed.

\paragraph{Definition.}
In meta-learning, we
collect a meta-training set $D_{\ML\text{-}\tr}=\{({D_{\tr}}_j\in\sI, h^*_j\in\sO)\}_{j=1}^{N_{\ML}}$, composed of $N_\ML$ i.i.d. samples from an unknown meta distribution $\sD_\ML$ on $\sI\times\sO$. We call $D_{\tr}$ a training set (meta input), $h^*$ a ``target'' base model (meta output), and $(D_{\tr}, h^*)$ a meta labeled example (task).  We call $g:\sI\mapsto\sO$ a meta model (meta mapping), which outputs a base model for a training set. For instance, in one-shot $C$-way learning for image classification, $D_{\tr}=\{(x_n ,y_n)\}_{n=1}^N$ contains $N$ labeled images, one for each of the $C$ classes (i.e., $N=C$). $h^*$ is a strong classifier trained using ample labeled images.

Meta-learning searches for a meta model $g$ given $D_{\ML\text{-}\tr}$, so that $g$ will work well on $(D_{\tr}, h^*)$ sampled from $\sD_\ML$. That is, $g$ should have a small meta generalization error $L_{\sD_{\ML}}(g)$ according to a meta loss
$l_\ML: \sO \times\sO \mapsto \R$
\begin{align}
L_{\sD_{\ML}}(g) = E_{(D_{\tr}, h^*)\sim\sD_{\ML}}[l_{\ML}(g(D_{\tr}), h^*)].  \label{e_meta_test_error}
\end{align}
In one-shot learning, we can view $g$ as a predictor of (strong) classifiers given small training sets.

\paragraph{Algorithm.}
To this end, we can follow supervised learning to construct a meta hypothesis set $\sG=\{g\}$ of candidate meta models and design an algorithm $B_\sG$ to search $\hat{g}$ from $\sG$ by learning from $D_{\ML\text{-}\tr}$.
We denote $\hat{g} = B_\sG(D_{\ML\text{-}\tr})$. We can apply ERM
\begin{align}
B_\sG(D_{\ML\text{-}\tr}) & = \arg\min_{g\in\sG} \cfrac{1}{N_\ML}\sum_{j=1}^{N_\ML} l_\ML(g({D_\tr}_j), h^*_j) \nonumber\\ 
& = \argmin_{g\in\sG} L_{S_{\ML}}(g),\label{e_meta_ERM}
\end{align}
where $L_{S_{\ML}}$ is the meta training error. In practice, we design a set of $(\sG$, $B_\sG)$ pairs and select the best pair using a held-out meta-validation set $D_{\ML\text{-}\val} = \{({D_{\tr}}_m,h^*_m)\}_{m=1}^{M_{\ML}}$ sampled i.i.d. from $\sD_\ML$. This is called meta model selection (meta-validation).

\paragraph{Discussion.} The meta model $g$ and the supervised learning algorithm $A_\sH$ (cf. \autoref{e_RLM}) have the same forms of inputs and outputs. Thus, $g$ can be seen as a (supervised) learning algorithm which may involve an optimization process\footnote{Indeed, \citet{metz2018learning} point out that many meta-learning algorithms consist of two levels of learning, in which $g$ is applied at the inner loop while our $B_\sG$ is applied at the outer loop.}. Nevertheless, the generalizability of $g$ and $A_\sH$ are drastically different. $A_\sH$ is designed or selected by model selection specifically for $\sD$, while $g$ is learned from $D_\ML$ for the purpose of generalizing to tasks sampled from $\sD_\ML$.

Moreover, viewing $(D_\tr, h^*)$ as a meta labeled example enables $h^*$ to be disentangled from $D_\tr$; \ie, $h^*$ can be flexibly defined to provide supervision for various applications, and is not necessarily a base model that performs on data sampled exactly from where $D_\tr$ is sampled. This notion broadens the applicability of meta-learning --- \eg, to domain generalization where $h^*$ is for a different domain, or to unsupervised (feature) learning where $D_\tr$ contains only unlabeled examples $\{x_n\}_{n=1}^N$. See \autoref{S_principled} for more details.

\paragraph{Comparisons to previous work.}
Viewing meta-learning as learning a mapping $g$ from $D_\tr$ to $h$ has been studied in \citep{thrun1998learning,baxter2000model,maurer2005algorithmic}. However, they did not treat $(D_\tr, h^*)$ pairs as meta-training data (cf. \autoref{e_meta_test_error} and \autoref{e_meta_ERM}) to embody the flexibility in designing $h^*$. In contrast, they mainly analyzed the case where $g(D_\tr)$ is applied to data from the same distribution where $D_\tr$ is sampled: i.e., each task is a supervised learning problem and the meta-training objective is to minimize the training loss on $D_\tr$ across multiple tasks with generalization guarantees. In other words, they treated $(D_\tr, D_\tr)$ pairs as meta-training data. Their studies may not be applied to recent meta-learning, where validation data is explicitly used to optimize the meta model (cf. \autoref{e_meta_few}), and to applications beyond supervised learning, such as domain generalization and continual learning~\citep{kaiser2017learning}, where $h^*$ is designed for another data distribution. \citet{finn2018learning, larochelle2018few} considered $(D_\tr, D_\val)$ pairs but do not explicitly allow $D_\val$ to be sampled from a different distribution. Our work is inspired by \citep{wang2016learning,wang2017learning,wang2018learning2} which explicitly defined $h^*$ for few-shot learning.

\subsection{Generalizability of learned meta models}
\label{S_general}
This unifying view enables transferring experiences of learning a base model
to learning a meta model. For example, increasing the size of meta-training set or minimizing the domain shift between $\sD_\ML$ and where the novel tasks will be sampled~\citep{ben2010theory,gong2012geodesic} should improve the generalization ability of the learned meta model $\hat{g}$. We empirically verify that these experiences are applicable in \autoref{S_exp}.

We can further apply theoretical analysis of supervised learning. Suppose $l_\ML$ is bounded and $|\sG|$ is finite, the Chernoff bound implies that $\sG$ is agnostic PAC learnable using ERM~\citep{shalev2014understanding}. This has indeed been exploited in~\citep{garg2018supervising} to derive a meta-learning bound, but only for meta unsupervised learning. 

\subsection{A principled way to apply meta-learning}
\label{S_principled}

\begin{figure}
	\centering
	\includegraphics[width=0.9\linewidth]{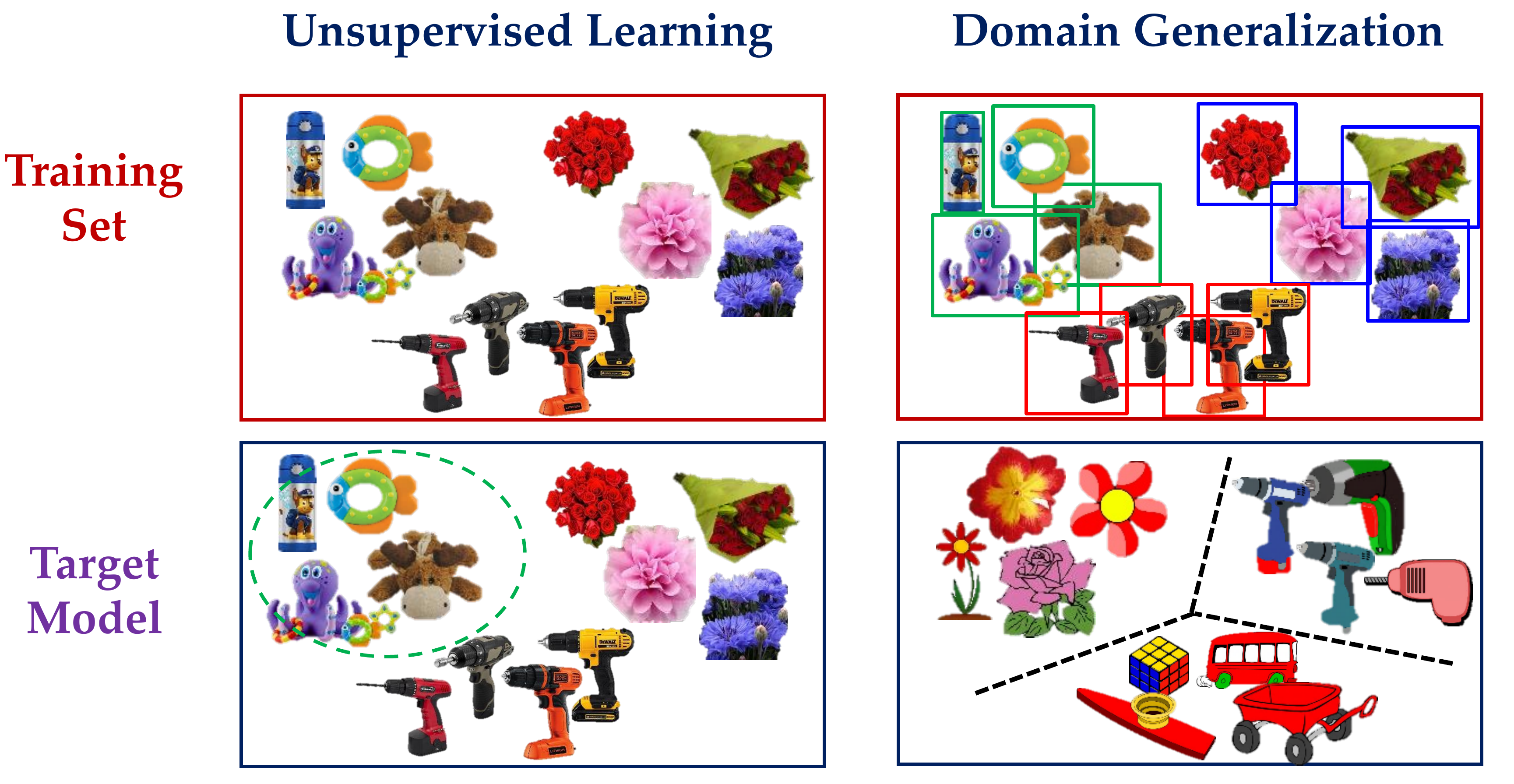}
	\vskip -10pt
	\caption{\small \textbf{Meta labeled examples.} A labeled image is marked by a colored box that indicate its label. The target model in unsupervised learning is a feature extractor that facilitates clustering or classification. The target model for domain generalization is a classifier for a different domain.}
\label{fig:application}
\vskip -10pt
\end{figure}

This unifying view clearly defines the components of meta-learning: (a) meta labeled example $(D_\tr, h^*)$ and meta loss $l_\ML$; (b) meta hypothesis set $\sG=\{g\}$ and meta algorithm $B_\sG$. 
Applying meta-learning to an area thus requires defining or designing them accordingly. We discuss two examples below, and \autoref{fig:application} gives an illustration. Please see suppl. for additional examples. 

\paragraph{Unsupervised learning.} Let us consider learning a feature extractor from unlabeled data to benefit downstream applications~\citep{boney2018semi,metz2018learning,yu2018one}. Here, $D_\tr=\{x_n\}_{n=1}^N$, $h^*$ is the target feature extractor, and $l_\ML$ measures the performance gap on the downstream application. $\sG$ is a set of objective functions for unsupervised learning; $B_\sG$ can be ERM by stochastic gradient descent (SGD). 

\paragraph{Domain generalization.}
The goal is to train a base model on source domains (SD) and apply it to target domains (TD) without fine-tuning. Here, $D_\tr=\{(x_n,y_n)\}_{n=1}^N$ is the labeled data from SD, while $h^*$ is the target base model that works well on TD. $l_\ML$ is the performance gap on TD. $g$ can be a model predictor or a learning process aware of the domain shifts~\citep{li2018learning,balaji2018metareg,li2019feature}. $B_\sG$ can be ERM by SGD.

\begin{algorithm}[t]
	\caption{\small \textbf{ meta-KNN} (cf. \autoref{e_meta_ERM})}	\label{a_meta_KNN}
	\textbf{Required:} $D_{\ML\text{-}\tr}$: a meta-training set \\ 
	\hspace{51pt} $\alpha$, $\beta$, $K$: hyper-parameters \\
	\hspace{51pt} $B_\sG$: an iterative meta-learning Algo., \\ 
	\hspace{71pt} with $\hat{g} = B_\sG(D_{\ML\text{-}\tr})$ converged. \\
	\textbf{Meta input:} A novel test task $D_\tr^\text{test}$\\
	1. Search $K$NN($D_\tr^\text{test}$): $K$NN tasks in $D_{\ML\text{-}\tr}$\\
	2. Copy $\tilde{g}$ from $\hat{g}$. Fine-tune $\tilde{g}$ on $K$NN($D_\tr^\text{test}$), by applying $B_\sG$ for $\beta$ extra epochs with step size $\alpha$\\
	\textbf{Meta output:} $\tilde{g}(D_\tr^\text{test})$
\end{algorithm}

\subsection{Supervised learning techniques}

Our unifying view enables transferring techniques of supervised learning to meta-learning with minimum adjustment. In our experiment, we adapt widely-used techniques for improving generalization abilities or facilitating optimization, including data augmentation, ensemble methods~\citep{breiman1996bagging,zhou2012ensemble,dietterich2000ensemble}, joint training (e.g., multi-task learning~\citep{argyriou2007multi}), and pre-training~\citep{yosinski2014transferable} to meta-learning.

Besides, it is well-known that non-parametric models~\citep{weinberger2009distance} are able to capture local and heterogeneous structures in data. Contrast to the fact that most existing meta models are parametric, we present a procedure called meta-KNN (\autoref{a_meta_KNN}), inspired by SVM-KNN~\citep{zhang2006svm,chao2013facial}, to build a non-parametric meta model. Meta-KNN begins with training a conventional meta model $\hat{g}$ on the meta-training set $D_{\ML\text{-}\tr}$. During meta-testing, given a test task $D_\tr^\text{test}$, meta-KNN then searches for its $K$ nearest neighbor (KNN) tasks from $D_{\ML\text{-}\tr}$, and fine-tune $\hat{g}$ using $B_G$ to minimize the meta-training loss computed only on those KNN tasks. The fine-tuned $\hat{g}$ (\ie, $\tilde{g}$) is then applied to the test task $D_\tr^\text{test}$. We empirically show its superior performance on meta-learning from heterogeneous domains.

\section{Case Study: Few-Shot Learning}
\label{S_appl}
\vspace{-5pt}
We present a case study on how to apply our meta-learning framework to few-shot learning. The goal of few-shot learning is to quickly build a model for a novel task; i.e., with minimum training time and training data. Specifically, we focus on one-shot $C$-way learning for image classification.

\subsection{Meta labeled examples and meta losses} The training set $D_\tr$ contains one labeled image for each $C$ classes; i.e., $D_\tr=\{(x_n,y_n)\}_{n=1}^{N}$ ($C=N$ in this case). The target model $h^*$ is a $C$-way classifier and the loss $l_\ML(h, h^*)$ measures how different $h$ compared to $h^*$. We present two examples of $l_\ML(h, h^*)$ as follows.

\paragraph{Meta losses in the model space.}
$h^*$ is the target classifier trained on a larger training set. Let $h$ be parameterized by $\theta$, $l_\ML(h, h')=\|\theta-\theta'\|^2_2$~\citep{wang2016learning, wang2017learning,gui2018few}.
	
\paragraph{Meta losses from example losses.} Given a validation set  $D_\val=\{(x_m,y_m)\}_{m=1}^{M}$ sampled in the same way as $D_\tr$, a reasonable choice of $l_\ML(h, h^*)$ is $|L_V(h)-L_V(h^*)|$, where $L_V$  is the validation error defined in \autoref{e_val}. Suppose $h^*$ minimizes $L_V$, then $l_\ML(h, h^*)$ is equivalent to $L_V(h)$. In other words, we replace $h^*$ and $l_\ML$ by $D_\val$ and $l$. The meta-training set $D_{\ML\text{-}\tr}$ thus becomes $\{({D_{\tr}}_j, {D_{\val}}_j)\}_{j=1}^{N_{\ML}}$ and ERM in \autoref{e_meta_ERM} can be re-written accordingly as (with constants ignored)
\begin{align}
\vspace{-5pt}
\hat{g} & = \argmin_{g\in\sG} \sum_{j=1}^{N_\ML} {L_V}_j(g({D_{\tr}}_j)) \nonumber \\ & = \argmin_{g\in\sG} \sum_{j=1}^{N_\ML} \sum_{m=1}^{M} l(g({D_\tr}_j)(x_{jm}), y_{jm}). \label{e_meta_few}
\vspace{-5pt}
\end{align}
${L_V}_j$ is the validation loss on ${D_\val}_j=\{(x_{jm},y_{jm})\}_{m=1}^{M}$. \autoref{e_meta_few} has been applied in many few-shot learning algorithms~\citep{hariharan2017low,ravi2017optimization,gidaris2018dynamic,qiao2018few,snell2017prototypical,sung2018learning,bertinetto2016learning,vinyals2016matching} and we will focus on it in the next subsections.

\subsection{Meta models and meta algorithms} We discuss two exemplar designs of $g$: one views $g$ as a supervised learning algorithm~\citep{finn2017model} and the other views $g$ as a feed-forward model predictor~\citep{snell2017prototypical}. Both algorithms are used in our experiments. 

\paragraph{Meta models as learning algorithms.}
Since $g$ outputs a classifier $h$ given $D_\tr$, it can be seen as a learning algorithm. Suppose $h$ is parameterized by $\theta$, let us apply ERM
\begin{align}
\vspace{-5pt}
g(D_\tr) =\arg\min_\theta L_S(\theta) =  \arg\min_\theta \cfrac{1}{N}\sum_{n=1}^{N} l(h_\theta(x_n), y_n).\nonumber
\vspace{-5pt}
\end{align}
Since $D_\tr$ is small, ERM may suffer over-fitting. One solution is to apply an iterative optimizer (e.g., gradient descent (GD)) with early stopping and a carefully chose initialization $\psi$~\citep{finn2017model}: early stopping limits the hypothesis set while $\psi$ prevents under-fitting. Suppose $L_S(\theta)$ is differentiable w.r.t. $\theta$, $g$ with one-step GD of a step size $\alpha$, initialized at $\psi$, is
\begin{align}
g_\psi(D_\tr) = \hat{\theta} = \psi - \alpha \times \nabla_\theta L_S(\psi).\label{e_MAML_inner}
\end{align}
The meta hypothesis set $\sG$ thus becomes $\{g_\psi\}$ with different initializations. To search $g_{\hat{\psi}}$, we apply ERM but on the meta-training set $D_{\ML\text{-}\tr}$ (cf. \autoref{e_meta_few}) 
\begin{align}
\hat{\psi} = & \argmin_\psi \sum_{j=1}^{N_\ML} \sum_{m=1}^{M} l(g_\psi({D_\tr}_j)(x_{jm}), y_{jm}). \label{e_MAML}
\end{align}
If we apply SGD for optimization then this is the one-step MAML~\citep{finn2017model}. In other words, MAML and its variants~\citep{lee2018meta,finn2018meta,li2017meta,rusu2019meta} fit in our framework.

\paragraph{Meta models as model predictors.}
Alternatively, we can view $g(D_\tr)$ as a model predictor. For example, in Prototypical Network (ProtoNet)~\citep{snell2017prototypical}
\begin{align}
g_\psi(D_\tr)(x) = \argmax_c \exp(-\|\psi(x) - \psi(x_{c})\|_2^2), \label{e_ProtoNet}
\end{align}
where $x_{c}$ is the image of class $c$, and $\psi$ is a feature extractor.  \citet{snell2017prototypical} learns $\psi$ via ERM (\autoref{e_meta_few}) and fits in our framework. Similar approaches are~\citep{qiao2018few,sung2018learning,wang2016learning,vinyals2016matching,wang2017learning}.

\subsection{Collecting meta-training sets}
After identifying the form of meta labeled example, which is $(D_\tr, D_\val)$ for few-shot learning, the next step is to collect the meta-training set. While collecting $(D_\tr, D_\val)$ one-by-one seems standard, it might be inefficient. For image classification it is easier to collect images class-by-class to first create a pool (named meta-train-pool), which contains many classes and each has many examples.
We then synthesize $(D_\tr, D_\val)$ from the pool. This is the general setting for  few-shot image classification. There will be three pools of disjoint classes: meta-train pool, meta-val-pool, meta-test-pool.

\section{Related Work}
\label{S_related}
There are excellent overviews and surveys of meta-learning~\citep{thrun1998learning,vilalta2002perspective,lemke2015metalearning,vanschoren2018meta,finn2018learning,metz2018learning}. 
There are also theoretical analysis~\citep{baxter2000model,maurer2016benefit,maurer2005algorithmic}. However, the flexibility to design $h^*$ is not considered.
\citet{franceschi2017bridge} connected hyperparameter tuning and meta-learning, which aligns with the comparison of $A_\sH$ and $g$ in \autoref{S_meta_idea}. \citet{garg2018supervising} also related meta-learning to supervised learning but only for meta-unsupervised learning.
Meta-learning has also been applied to reinforcement learning and imitation learning~\citep{stadie2018importance,frans2017meta,wang2016RL,duan2016rl,duan2017one,finn2017one,yu2018one},
optimization~\citep{andrychowicz2016learning,wichrowska2017learned, li2017learning,bello2017neural},
recommendation systems~\citep{vartak2017meta}, data augmentation~\citep{ratner2017learning}, natural language processing~\citep{huang2018natural}, architecture search~\citep{elsken2018neural}, continual learning~\citep{riemer2019learning,kaiser2017learning,al2018continuous,clavera2019learning}, transfer and multi-task learning~\citep{ying2018transfer,zhang2018learning}, active learning~\citep{ravi2018meta,sharma2018learning, bachman2017learning,ravi2018meta,sharma2018learning, pang2018meta}, and teaching~\citep{fan2018learning}. Some algorithms aim for versatile purposes~\citep{,mishra2018simple,munkhdalai2017meta,ritter2018been,finn2017model,santoro2016meta,nichol2018first}.

\section{Experiments}
\label{S_exp}
\vspace{-5pt}
We validate the advantages of our unifying view by investigating three aspects of few-shot learning: (a) whether the number of meta-examples influences the generalization; (b) whether supervised learning techniques are applicable to meta-learning; (c) whether our unifying view can facilitate new applications and algorithm design. We first describe the setups. 

\paragraph{Datasets.} The {\it Mini}ImageNet dataset~\citep{vinyals2016matching} is a subset of ImageNet~\citep{RussakovskyDSKS15ImageNet} and is widely-used in few-shot learning. There are 100 classes and 600 examples per class. We split the datasets following~\citep{ravi2017optimization}: there are 64, 16, 20 classes for meta-train-pool, meta-val-pool, and meta-test-pool. We also consider a challenging case where the splits are 30, 30, 40 classes, in which the diversity of meta-training examples is limited. We call the former standard split (SS) and the later challenging split (CS).
We further consider two set of datasets. To enlarge the heterogeneity among tasks, we synthesize a ``Heterogeneous'' dataset from five fine-grained datasets, namely AirCraft~\citep{maji13finegrained}, Car-196~\citep{Krause3DRR2013},  Caltech-UCSD Birds (CUB) 200-2011~\citep{WahCUB_200_2011}, Stanford  Dog~\citep{KhoslaFGVC2011}, and Indoor~\citep{Quattoni2009Recognizing}. We sample 60 classes from each dataset, and equally split them into meta-train-pool, meta-val-pool, and meta-test-pool.
To investigate the applicability of meta-learning, we use the Office-Home datasets~\citep{venkateswara2017Deep} in a  domain generalization problem. There are 65 classes and 4 domains of images per class. We test two domains, ``Clipart'' and ``Product.'' See suppl. for more details.

\paragraph{Meta examples (tasks) and evaluation protocols.} We focus on the 1-shot 5-way tasks, unless stated otherwise. We follow~\citep{rusu2019meta} to draw 10,000 tasks $(D_\tr, D_\val)$ from meta-test-pool and there are 15 validation images per class in a task, except for \autoref{tab:non_para}.
We report the mean accuracy.
We found the 95\% confidence interval to be consistently within $[0.001, 0.004]$ and thus omit it for brevity.

\paragraph{Baselines.} We investigate Prototypical Network~(ProtoNet)~\citep{snell2017prototypical}, Matching Network (MatchNet)~\citep{vinyals2016matching}, and Model Agnostic Meta-Learning (MAML) \citep{finn2017model}. 
We implement the algorithms and use the same $C$-way setting in meta-training and meta-test for consistency: we disregard the trick~\citep{snell2017prototypical} that trains with $30$-way tasks but tests with $5$-way tasks. 
We use the standard 4-layer ConvNets as the backbone~\citep{vinyals2016matching,snell2017prototypical} and resize images to $84 \times 84$ following~\citep{vinyals2016matching}. More details, together with the results using ResNet~\citep{he2016deep}, are in suppl.

\subsection{Generalization analysis}
\label{S_GA}
\begin{figure}
	\centering
	\begin{minipage}[h]{0.325\linewidth}
		\centering \includegraphics[width=\linewidth]{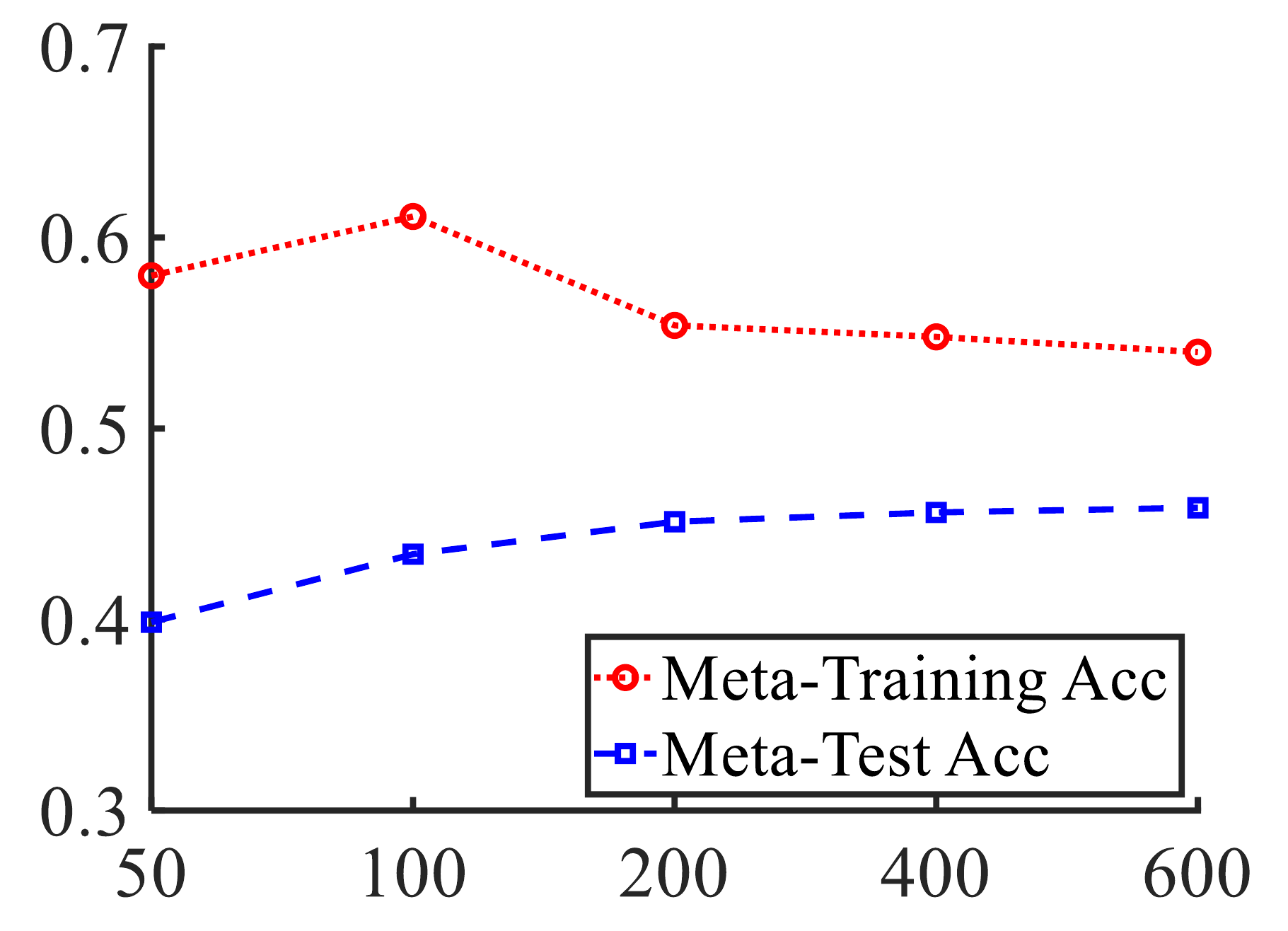}\\
		\vskip -3pt
		\mbox{\small ({a}) {MAML}}
	\end{minipage}
	\begin{minipage}[h]{0.325\linewidth}
		\centering
		\includegraphics[width=\linewidth]{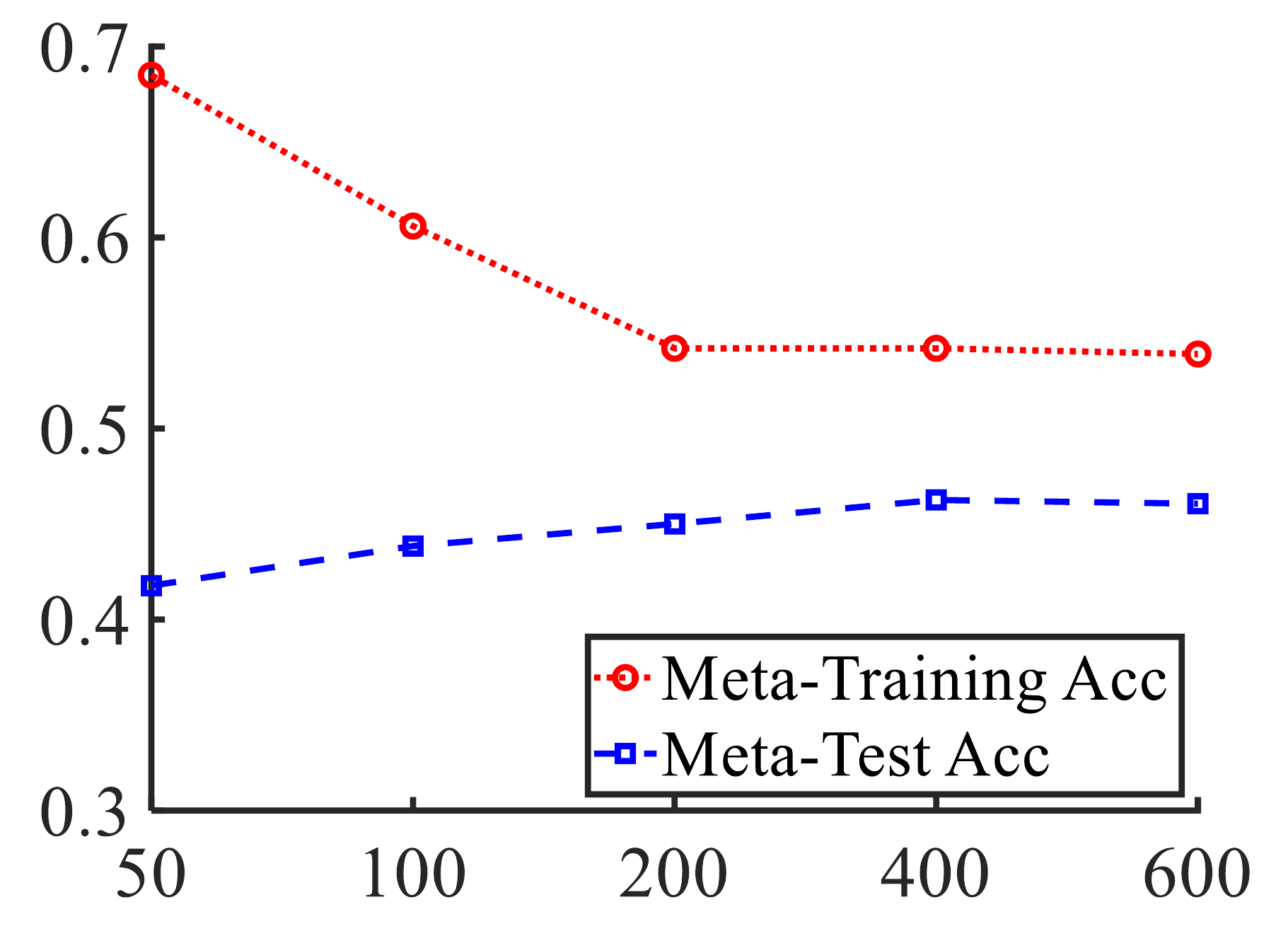}\\
		\vskip -3pt
		\mbox{\small ({b}) {ProtoNet}}
	\end{minipage}
	\begin{minipage}[h]{0.325\linewidth}
		\centering
		\includegraphics[width=\linewidth]{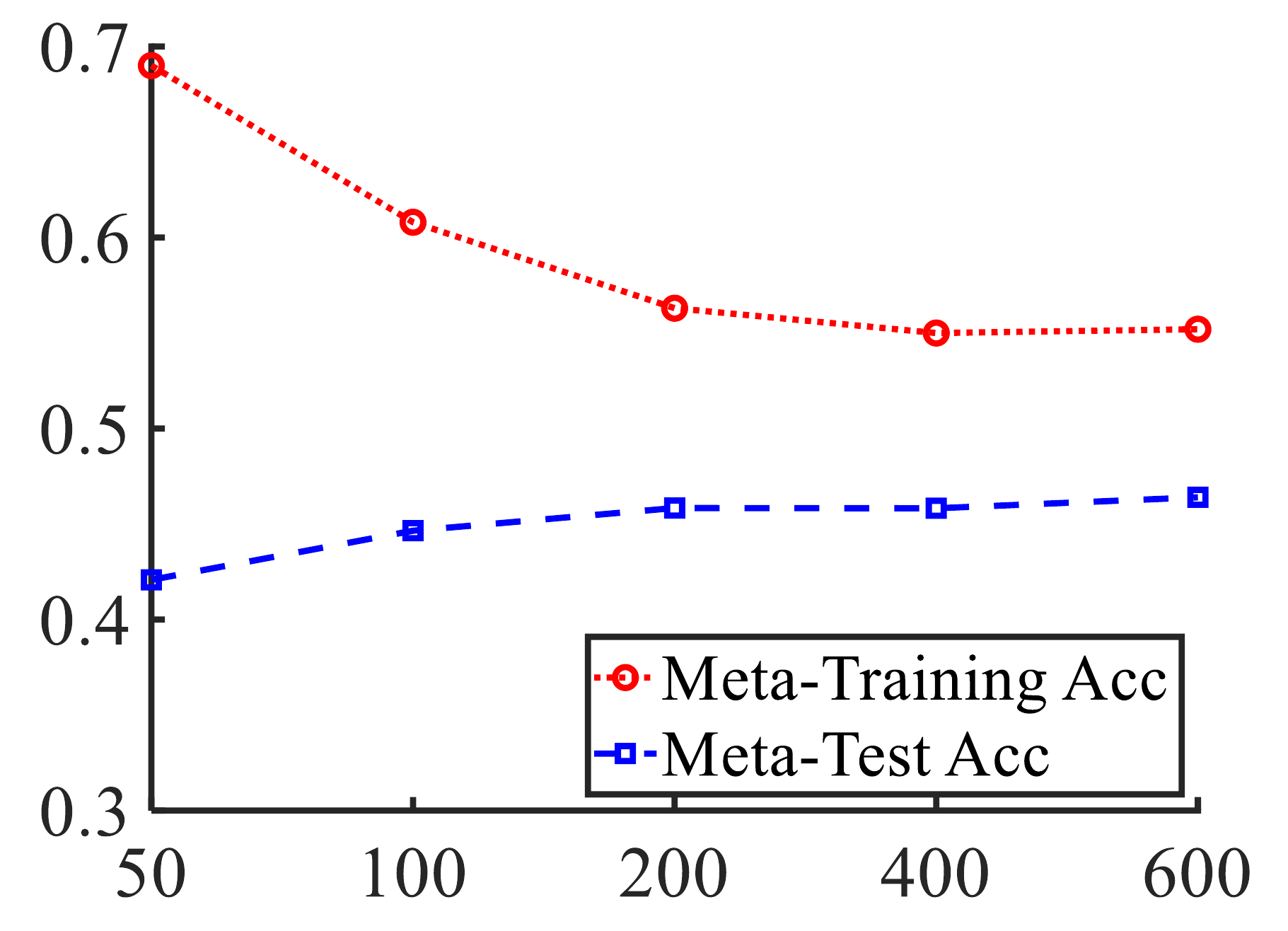}\\
		\vskip -3pt
		\mbox{\small ({c}) {MatchNet}}
	\end{minipage}
	\vskip -5pt
	\caption{\small 1-shot 5-way accuracy on {\it Mini}ImageNet (SS). We show meta-training (red) and meta-test (blue) accuracy. We vary the numbers of instances per class in meta-train-pool to control the number of meta-training examples. All algorithms suffer over-fitting when the number gets small. }
	\label{fig:generalization}
	\vskip -10pt
\end{figure}

Following our supervised view of meta-learning, the generalization ability of the learned meta model should be affected by the ``effective'' number of meta training examples; i.e., the number of different tasks. We validate the influence with various meta-train-pool configurations on {\it Mini}ImageNet (SS). We keep all the 64 classes in meta-train-pool to keep the meta distribution intact. Instead, we change the number of instances in each class from 600 to 50 to construct different meta-train-pools. When there are limited instances per class in the meta-train-pool, the number of unique tasks is constrained as well. We also evaluate the case of fewer classes in suppl. 
\autoref{fig:generalization} shows the change of few-shot accuracy with different meta-train-pools. The trend follows the supervised view: the more meta-training examples (few-shot tasks) are provided, the better generalization is achieved. Specifically, learning with 50 instances per class significantly over-fits.

\subsection{Supervised learning techniques}
\label{S_SLML}
We investigate five techniques: joint (multi-objective) learning, pre-training, bagging, data augmentation, and non-parametric methods. We experiment on {\it Mini}ImageNet (SS), unless stated otherwise.

\paragraph{Multi-objective learning.} In addition to the original objective (i.e., meta-training loss), learning jointly with related objectives has shown promising results in supervised learning: it serves as a data-dependent regularization to improve generalization. Here we add another objective: a 64-way classification with cross-entropy loss over all classes in meta-train-pool. This classifier shares weights with the meta model, except for the last layer. In other words, the shared sub-network should jointly master two tasks: predicting good $C$-way classifiers and extracting discriminative features for 64-way classification.
\citet{oreshkin2018tadam} explored this idea, yet we provide more insights in \autoref{tab:joint_train}. 

\paragraph{Pre-training.} It is well-known that a good model initialization significantly facilitates the model optimization for down-stream task~\citep{erhan2010does,bengio2007greedy}. Specifically, supervised pre-training~\citep{yosinski2014transferable} on a large labeled dataset has been prevalent in many applications. While the standard setup in few-shot learning trains the meta model from scratch on the sampled meta labeled examples,
we investigate training a 64-way classifier with cross-entropy loss at first, and use it to initialize the meta model. Such a strategy is explored in~\citep{chen2019closer,qiao2018few,rusu2019meta,li2018network}, yet we provide more insights as follows\footnote{We note that~\citet{triantafillou2019meta} also provided detailed studies on pre-training in their latest version.}.

The results by applying the two techniques are shown in \autoref{tab:joint_train}. We list both the meta-training and meta-test accuracy. Both techniques consistently improve the test accuracy for all three algorithms. \emph{However, their underlying influences are different.} Pre-training achieves the highest meta-training accuracy, justifying its effectiveness to facilitate optimization. Multi-objective learning does not increase the meta-training accuracy much but improves the meta-test accuracy, verifying its ability to improve generalization. In summary, we show that well-known supervised learning techniques are applicable to meta-learning in the same manner they benefit supervised learning.

\begin{table*}[t]
	\centering
	\small
	\tabcolsep 3pt
	\caption{\small Multi-objective learning and pre-training for 1-shot 5-way classification on {\it Mini}ImageNet (SS).
	}
	\vskip -7pt	
	\begin{tabular}{l|c|c|c|c|c|c|c|c|c}
		\hline
		& \multicolumn{3}{c|}{MAML} & \multicolumn{3}{c|}{ProtoNet} & \multicolumn{3}{c}{MatchNet} \\ \cline{2-10}
		& Scratch & Pre-Train & Multi-Obj. & Scratch & Pre-Train & Multi-Obj. & Scratch & Pre-Train & Multi-Obj. \\ \hline
		Train & 0.540 & 0.576 & 0.554 & 0.539 & 0.602 & 0.537 & 0.547 & 0.662 & 0.598\\
		Test & 0.459 & 0.478 & 0.470 & 0.461 & 0.500 & 0.484 & 0.463 & 0.500 & 0.485\\ \hline
	\end{tabular}
	\label{tab:joint_train}
	\vskip-5pt
\end{table*}

\paragraph{Ensemble methods.} Ensemble methods leverage the diversity among a set of basic models to construct a robust summarized model. It has been comprehensively verified in supervised learning. We apply a simple ensemble method bagging~\citep{breiman1996bagging} to meta-learning, which reduces the model variances by majority voting over many classifiers. We generate diverse meta models by learning from different meta-training sets. Concretely, we sample 10 different sub meta-train-pools from the original one: each sub-pool contain 48 classes. Then we train 10 basic meta models over those sub meta-train-pools and ensemble the results for final tests.

\begin{table*}[t]
	\centering
	\small
	\tabcolsep 2.5pt
	\caption{\small Bagging for 1/5-shot 5-way classification on {\it Mini}ImageNet (SS). Single: no bagging. Average: average accuracy of basic meta models. The best result of each shot-algorithm pair is in bold font.}
\vskip-7pt
\begin{tabular}{c|r|c|cc|c|cc|c|cc}
	\hline
	\multicolumn{2}{c|}{} & \multicolumn{3}{c|}{MAML} & \multicolumn{3}{c|}{ProtoNet} & \multicolumn{3}{c}{MatchNet} \\ \cline{3-11}
	\multicolumn{2}{c|}{} & Single & Average & Bagging &  Single & Average & Bagging &  Single & Average & Bagging \\ \hline
	\multirow{2}{*}{1-shot} & w/o Pre-Train   & 0.459 & 0.448 & 0.482 & 0.461 & 0.463 & 0.491 & 0.463 & 0.457 & 0.482 \\ 
	& w/ Pre-Train    & 0.478 & 0.439 & \textbf{0.498} & 0.500 & 0.492 & \textbf{0.526} & 0.500 & 0.495 & \textbf{0.525} \\ \hline
	\multirow{2}{*}{5-shot} & w/o Pre-Train  & 0.633 & 0.617 & 0.666 & 0.658 & 0.649 &  0.694 & 0.639 & 0.625 & 0.664\\ 
	& w/ Pre-Train   & 0.660 & 0.646 & \textbf{0.692} & 0.671 & 0.657 & \textbf{0.701} & 0.642 & 0.630 & \textbf{0.673}\\
	\hline
\end{tabular}
	\label{tab:bagging}
	\vskip -10pt
\end{table*}

The results are in \autoref{tab:bagging}. ``Single'' is the model trained on the original meta-train-pool, the same as in \autoref{tab:joint_train}.
The average performance of all 10 basic meta models are in the ``Average'' column. (Note that this is not ensemble yet.)
We investigate ``Bagging'' by combining the prediction probabilities of the 10 models.
The average performance of basic meta models is mostly worse than that of the meta model trained on the full 64-class meta-train-pool. But after we ensemble the basic models the performance improves notably, validating the effectiveness of ensemble for meta-learning. Moreover, we see that pre-training and bagging are compatible and lead to impressive 0.526/0.701 accuracy for 1/5-shot learning, which is on par with one state-of-the-art~\citep{qiao2018few} but with a cleaner meta-training procedure. In suppl. we also investigate this compatibility on CUB~\citep{WahCUB_200_2011} and {\it tiered}ImageNet~\citep{ren2018meta} and see similar trends of improvement. We also experiment with more advanced meta-learning algorithms~\citep{sung2018learning,ye2018learning} and backbones~\citep{he2016deep} in suppl.

\emph{Given that the above three techniques appear to be meta-model-agnostic by improving all three meta-learning models consistently, we focus on ProtoNet in the following experiments.}

\begin{table}
	\centering
	\small
	\caption{\small Test accuracy of 1-shot 30-way classification on {\it Mini}ImageNet (CS). $K$: cluster number to augment meta-train-pool,  meta-val-pool classes. $K=1$: no augmentation.}
	\vskip -7pt
	\begin{tabular}{cccc}
		\hline
		$K$ = 1 & $K$ = 2 & $K$ = 4 & $K$ = 8 \\
		\hline
		0.146 & 0.146 & 0.155 & 0.161 \\
		\hline
	\end{tabular}
	\vskip -5pt
	\label{tab:class_aug}
\end{table}

\paragraph{Data augmentation.} Increasing the number of training examples by data augmentation is a popular technique in supervised learning to improve generalization, especially when the training examples are not diverse enough. Here we adapt the idea to meta-learning. We consider a challenging yet more realistic case on {\it Mini}ImageNet (CS). 
Specifically, we re-split the dataset into 30, 30, 40 classes for meta-train-pool, meta-val-pool, and meta-test-pool, and focus on $1$-shot~$30$-way classification. Note that, this setting might break the applicability of meta-learning to few-shot learning: every time we will sample the same 30 classes from meta-train-pool, greatly limiting the diversity of meta labeled examples. However, this setting is indeed more realistic in practice according to the well-known long-tailed distribution~\citep{SudderthJ08,SalakhutdinovTT11,zhu2014capturing,wang2017learning}: there are more few-shot classes than many-shot classes.

We present a data augmentation strategy at the meta example (task) level, inspired by~\citep{hsu2018unsupervised}. Specifically, we perform K-means within each class to split a class into $K$ subcategories, resulting in $30\times K$ augmented classes in meta-train-pool (and meta-val-pool). 
The results are in \autoref{tab:class_aug}. We cluster each class in the meta-train-pool and meta-val-pool with $K={1, 2, 4, 8}$ subcategories; $1$ means no K-means. We observe a clear trend: the more classes we augment to increase task diversity, the higher accuracy we can achieve. More details and results are in suppl.

\begin{table}
	\centering
	\small
	\caption{\small Parametric (ProtoNet) vs. non-parametric models (ProtoNet-KNN). We show the test accuracy of 1-shot 5-way fine-grained classification on the Heterogeneous dataset.}
	\vskip -7pt
	\begin{tabular}{c|c}
		\hline
		ProtoNet & ProtoNet-KNN \\
		\hline
  0.372 &  0.386 \\
		\hline
	\end{tabular}
	\vskip -5pt
	\label{tab:non_para}
\end{table}

\paragraph{Non-parametric methods.} Non-parametric approaches are known to capture local and heterogeneous structures in data. Here we extend the methods to meta-learning by applying meta-KNN (cf. \autoref{a_meta_KNN}) to ProtoNet, named ProtoNet-KNN. 
ProtoNet-KNN first learns a normal ProtoNet $\hat{g}$ on the meta-training set $D_{\ML\text{-}\tr}$. $B_G$ is therefore SGD that minimizes the meta-training loss of ProtoNet (\ie, \autoref{e_meta_ERM} and \autoref{e_meta_few}). In meta-test, given a test few-shot task $D_\tr^\text{test}$, we search for its $K$ nearest neighbor (KNN) few-shot tasks from the meta-training set $D_{\ML\text{-}\tr}$. We characterize each task by its average features and use the Euclidean distance to measure task similarity.
We then fine-tune $\hat{g}$ using $B_G$ (\ie, SGD), but to minimize the meta-training loss computed only on those KNN tasks\footnote{We note that since at first $\hat{g}$ has been trained to minimize the meta-training loss on $D_{\ML\text{-}\tr}$, fine-tuning $\hat{g}$ on the ``entire'' meta-training set $D_{\ML\text{-}\tr}$ will not lead to further improvement.}. The fine-tuned $\hat{g}$ (\ie, $\tilde{g}$) is then applied to the test task $D_\tr^\text{test}$.
Here we set $K = 100$ and fine-tune $\hat{g}$ for only $1$ epoch, which takes $<400$ ms. See suppl. for details and further discussions.

We investigate ProtoNet-KNN on the Heterogeneous dataset described earlier in setups. 
We focus on few-shot fine-grained classification: a test task contains 5 classes from one specific dataset. In meta-training, we sample a fine-grained task randomly from a dataset to train a single ProtoNet. We expect the learned ProtoNet to be equipped with dataset-agnostic fine-grained discriminative knowledge. In meta-test, we then apply ProtoNet-KNN to adapt the ProtoNet to the neighbor tasks of a test task. To facilitate neighbor searching, we pre-sample 2,000 training tasks from $D_{\ML\text{-}\tr}$ (cf. \autoref{a_meta_KNN}).
\autoref{tab:non_para} shows the meta-test results (on 2,000 tasks). Non-parametric methods greatly improve the performance on heterogeneous tasks. See suppl. for additional details including the hyper-parameters.

\subsection{Domain generalization \& domain shifts}
\label{S_DGMD}
\begin{table}[t]
	\centering
	\small
	\tabcolsep 2pt
	\caption{\small Few-shot 5-way domain generalization on Office-Home with two domains: Clipart (C), Product (P). We denote the source/target domain by SD/TD.
	}
	\vskip -7pt
	\begin{tabular}{c|cc|cc|cc}
		\hline
		& \multicolumn{2}{c|}{Meta-train} & \multicolumn{2}{c|}{Meta-test} & \multicolumn{2}{c}{Test Acc}\\ \cline{2-7}
		Case & SD & TD & SD & TD & 1-Shot & 5-Shot \\ \hline
		I-1 & C     & C     & C     & C     & 0.341 & 0.477 \\ 
		I-2 & P     & C     & P     & C     & 0.296 & 0.350\\
		I-3 & C     & C     & P     & C     & 0.275 & 0.342\\
		I-4 & P     & P     & P     & C     & 0.264 & 0.283\\
		\hline
	\end{tabular}
	\label{tab:domain}
	\vskip -10pt
\end{table}

Existing few-shot learning assumes that the training and validation examples within a task are from the same domain. In practice it is sometime desirable to remove the assumption; e.g., we may want to learn a robot in a constraint environment, where examples are not diverse and are essentially few-shot, but apply it in the wild. Recall that our supervised view allows the training set and target model $h^*$ of a task to be defined according to the problem at hand. By defining the target model $h^*$ to be the one that will work well in a different domain, we can remove the same-domain assumption systematically. We study this idea and apply the trick in \autoref{S_appl} to replace the target model $h^*$ by a validation set collected in a different domain.

We experiment on the Off-Home dataset described earlier in setups, which is designed for domain adaptation. We focus on two domains: Clipart (C) and Product (P). There are 65 classes and we split them into 25 for meta-train-pool, 15 for meta-val-pool, and 25 for meta-test-pool. We work on \emph{few-shot 5-way} tasks in which a task contains a training set $D_\tr$ from one (source) domain and a validation set $D_\val$ from the other (target) domain. 
We design several scenarios to illustrate the difference between domain generalization and meta-domain shift. In domain generalization, a task $(D_\tr, D_\val)$ has its $D_\tr$ and $D_\val$ sampled from two different domains. In meta-domain shift, the tasks in the meta-training set and meta-test set are sampled from different meta distributions. 

\autoref{tab:domain} shows the results. Comparing Case I-1 and I-2, we see the challenge of domain generalization when the source and target domains within a task are different. We note that in each of these cases the distribution of the meta-training and meta-test examples are the same\footnote{That is, I-1 and I-2 follow the assumption in supervised learning that the meta-training tasks and meta-test tasks are sampled from the same meta distribution.}. Therefore, there is no meta-domain shift and the gap simply indicates the difficulty of tasks.

We further investigate meta-domain shifts by constructing meta-training/-test examples in a different way (Case I-3 and I-4).
The results are outperformed by Case I-2. (Note that all three cases consider the same meta-test examples.) We attribute this gap to meta-domain shifts (cf. \autoref{S_general}), which can potentially be resolved via domain adaptation by casting meta-learning as supervised learning.
\section{Conclusion}
\label{S_disc}
\vspace{-5pt}
We revisit and strengthen the connection between meta-learning and supervised learning, upon which applying meta-learning  becomes systematic; theoretical and empirical understandings of supervised learning can be applied to meta-learning. 
We further demonstrate that various supervised learning techniques can benefit meta-learning in the same ways they benefit supervised learning. We hope our studies to re-inform the community of such a valuable connection and inspire new algorithms, theories, and understandings of meta-learning.

\section*{Acknowledgments}
This research is supported in part by grants from the National Science Foundation (III-1618134, III-1526012, IIS-1149882, IIS-1724282, and TRIPODS-1740822), the Office of Naval Research DOD (N00014-17-1-2175), the Bill and Melinda Gates Foundation, and the Cornell Center for Materials Research with funding from the NSF MRSEC program (DMR-1719875). We are thankful for generous support by SAP America Inc., and Facebook. We also thank Jie Fu (MILA), Ming-Yu Liu (NVIDIA), Po-Hsuan Chen (Google Brain), and Yu-Chiang Frank Wang (Natl. Taiwan Univ.) for discussions.

{\small
\bibliographystyle{plainnat}
\bibliography{ML}}

\clearpage
\appendix
\begin{center}
	\textbf{\Large Supplementary Material}
\end{center}

We provide details omitted in the main text.

\begin{itemize}
	\item \autoref{S_S_principled}: More examples on applying meta-learning (\autoref{S_principled} of the main paper).
	\item \autoref{S_exp_setup}: Details of experimental setups (\autoref{S_exp} of the main paper).
	\item \autoref{S_S_exp}: Additional experimental results (\autoref{S_GA}, \autoref{S_SLML} and \autoref{S_DGMD} of the main paper).
\end{itemize}

\section{More Examples on Applying Meta-Learning}
\label{S_S_principled}
Our unifying view clearly indicates the components of meta-learning.
\begin{itemize}
	\item Meta labeled example $(D_\tr, h^*)$ and meta loss $l_\ML$
	\item Meta hypothesis set $\sG=\{g\}$ and meta algorithm $B_\sG$
\end{itemize}

Applying meta-learning to an area requires defining or designing them accordingly. We discuss more examples in the following subsections.

\subsection{Active learning}
\label{S_active}
\begin{figure}
	\centering
	\includegraphics[width=\linewidth]{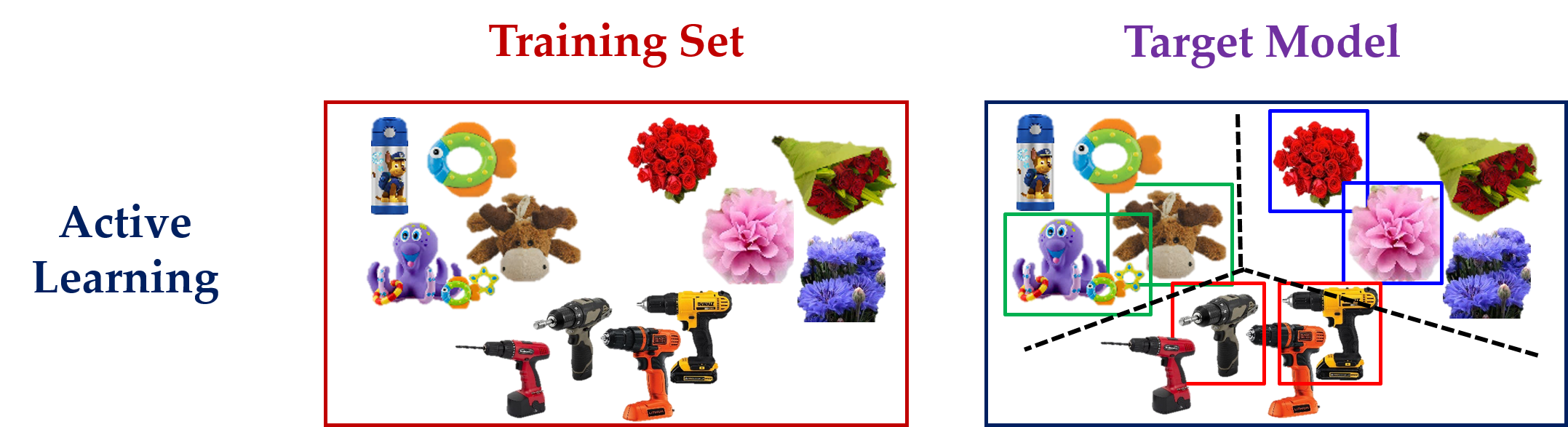}
	\caption{Illustration of active learning. A labeled image is surrounded by a colored box that indicate its label.  The target model in active learning is the classifier trained on the optimally queried images.}
	\label{fig:active_application}
\end{figure}

The goal is to minimize the data labeling effort by querying $N'$ informative examples. For classification, $D_\tr=\{x_n\}_{n=1}^{N}$ is an unlabeled set, $h^*$ is the model learned with the best $N'$ queried examples from $D_\tr$, and $l_\ML$ measures the performance gap on classification. $g$ is a learning process constrained to query labels for $N'$ examples by investigating $x$ and the learning progress~\citep{ravi2018meta,sharma2018learning, bachman2017learning}. $B_\sG$ can be ERM by SGD or by reinforcement learning algorithms~\citep{ravi2018meta,sharma2018learning}. An illustration of this procedure can be found in \autoref{fig:active_application}.

\subsection{Optimization}
\label{S_opt}
One focus is to search a better updating rule in iterative optimization. In this case, $D_\tr$ is an objective function $L_S(\theta)$ parameterized by $\theta$; e.g., the training loss. $h^*$ is the optimal solution of $L_S$ or of a related objective $L_V$; e.g., the validation error. $l_\ML$ is the gap of objective values. $g$ is an iterative optimization algorithm with a (meta) learnable updating rule. $B_\sG$ can be EMR by SGD~\citep{andrychowicz2016learning,wichrowska2017learned} or reinforcement learning algorithms~\citep{li2017learning,bello2017neural}.

\subsection{Imitation learning and reinforcement learning}
\label{S_Imit}
Imitation learning~\citep{stadie2018importance,frans2017meta,wang2016RL,duan2016rl,duan2017one,finn2017one,yu2018one} easily fits into our meta-learning framework, given its similarity to the supervised learning. For few-shot reinforcement learning~\citep{duan2017one,wang2016RL}, the $D_\tr$ is a trial of a few episodes from an MDP and $h^*$ is the target policy, which can be realized by maximizing the reward.

\subsection{Few-shot learning}
The line of learning generative models for few-shot learning~\citep{zhang2018metagan,wang2018low} can conceptually be thought of as learning a meta mapping from a (few-shot) training set to a target data generator (or distribution) that well describes many-shot examples. 

\section{Details of Experimental Setups}
\label{S_exp_setup}
\paragraph{Datasets.}
We synthesize the ``Heterogeneous'' dataset from five fine-grained classification datasets, namely AirCraft~\citep{maji13finegrained}, Car-196~\citep{Krause3DRR2013},  Caltech-UCSD Birds (CUB) 200-2011~\citep{WahCUB_200_2011}, Stanford  Dog~\citep{KhoslaFGVC2011}, and Indoor~\citep{Quattoni2009Recognizing}. We randomly sampled 60 classes with 50 images each from each of the 5 datasets, and equally split classes into meta-train-pool, meta-val-pool, and meta-test-pool. That is, there are 100 classes in meta-train-pool (same for the others), which includes 20 classes from each dataset.

We also use CUB~\citep{WahCUB_200_2011} alone to evaluate the few-shot learning performance. CUB contains a total of 11,788 images of birds over 200 species. The similarity between classes makes it a difficult classification task when given limited training examples. Following the setups of~\citep{TriantafillouZU17Few,chen2019closer} we use the provided bounding box of CUB to crop the center object of each image, and split the 200 classes into 100/50/50 for meta-train/val/test-pool.

\begin{figure*}[t]
	\begin{center}
	\begin{minipage}[h]{4.2cm}
		\centering \includegraphics[width=4.2cm ]{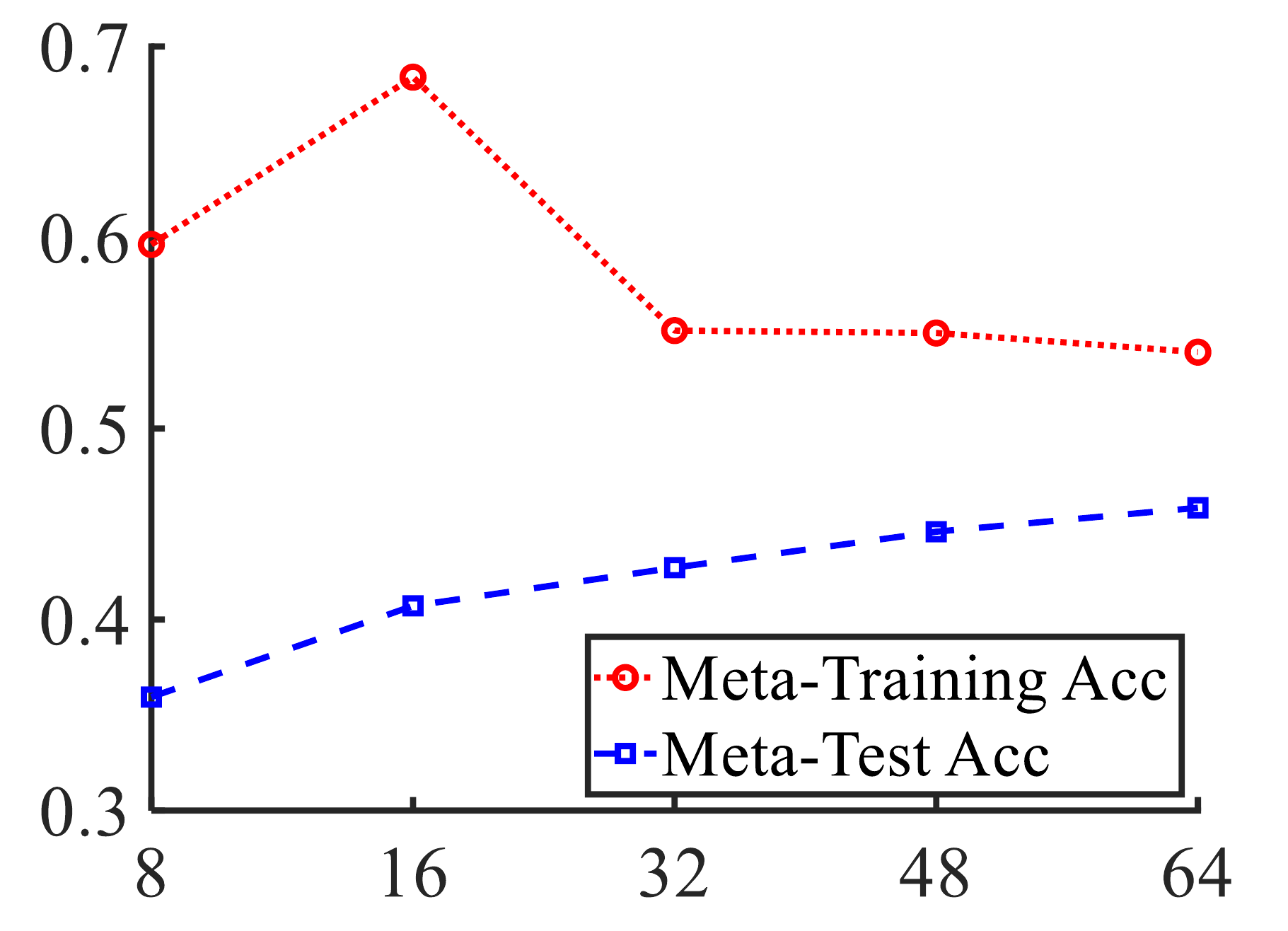}\\
		\mbox{\small ({\it a}) {MAML}}
	\end{minipage}
	\begin{minipage}[h]{4.2cm}
		\centering
		\includegraphics[width=4.2cm ]{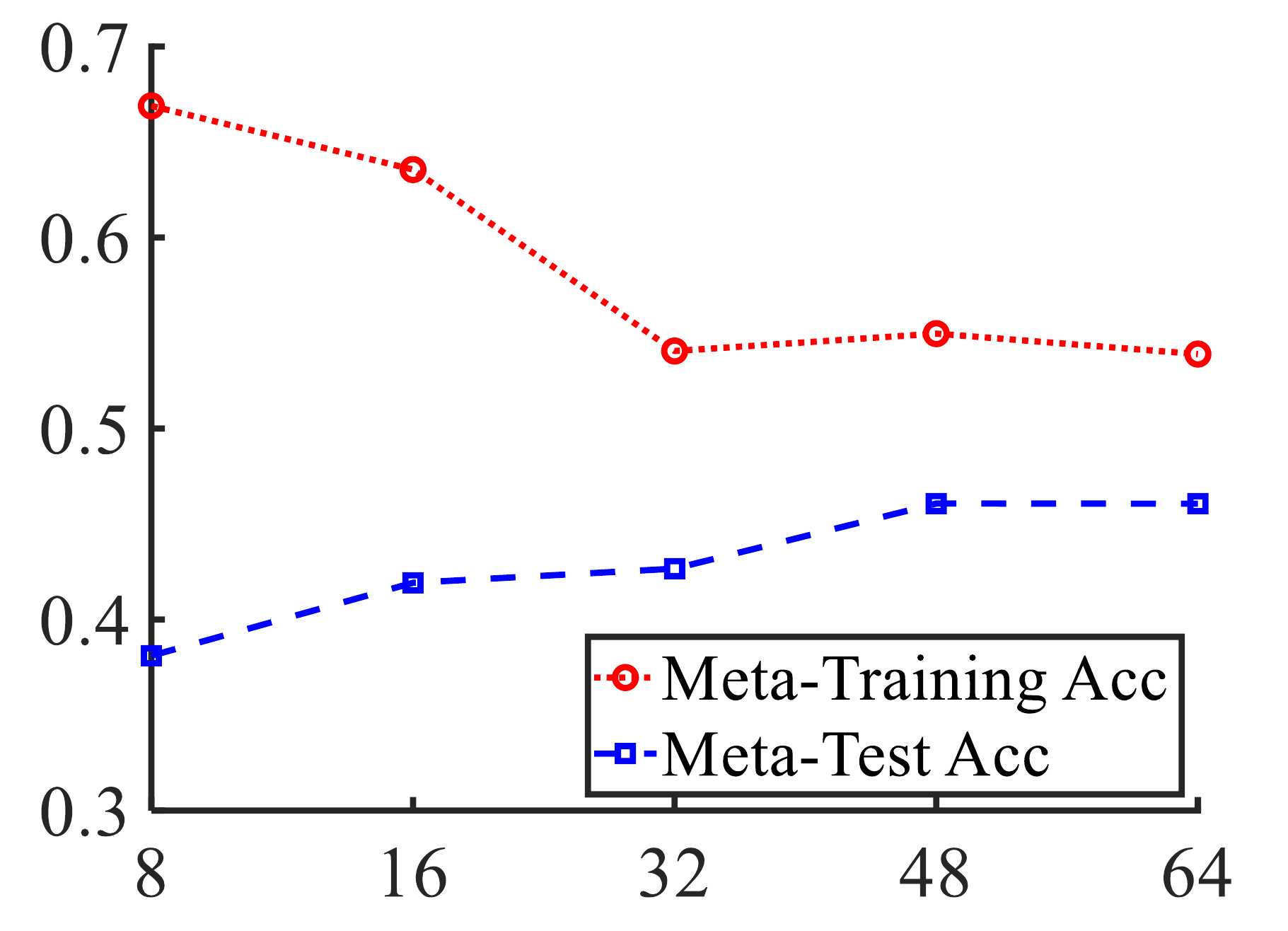}\\
		\mbox{\small ({\it b}) {ProtoNet}}
	\end{minipage}
	\begin{minipage}[h]{4.2cm}
		\centering
		\includegraphics[width=4.2cm ]{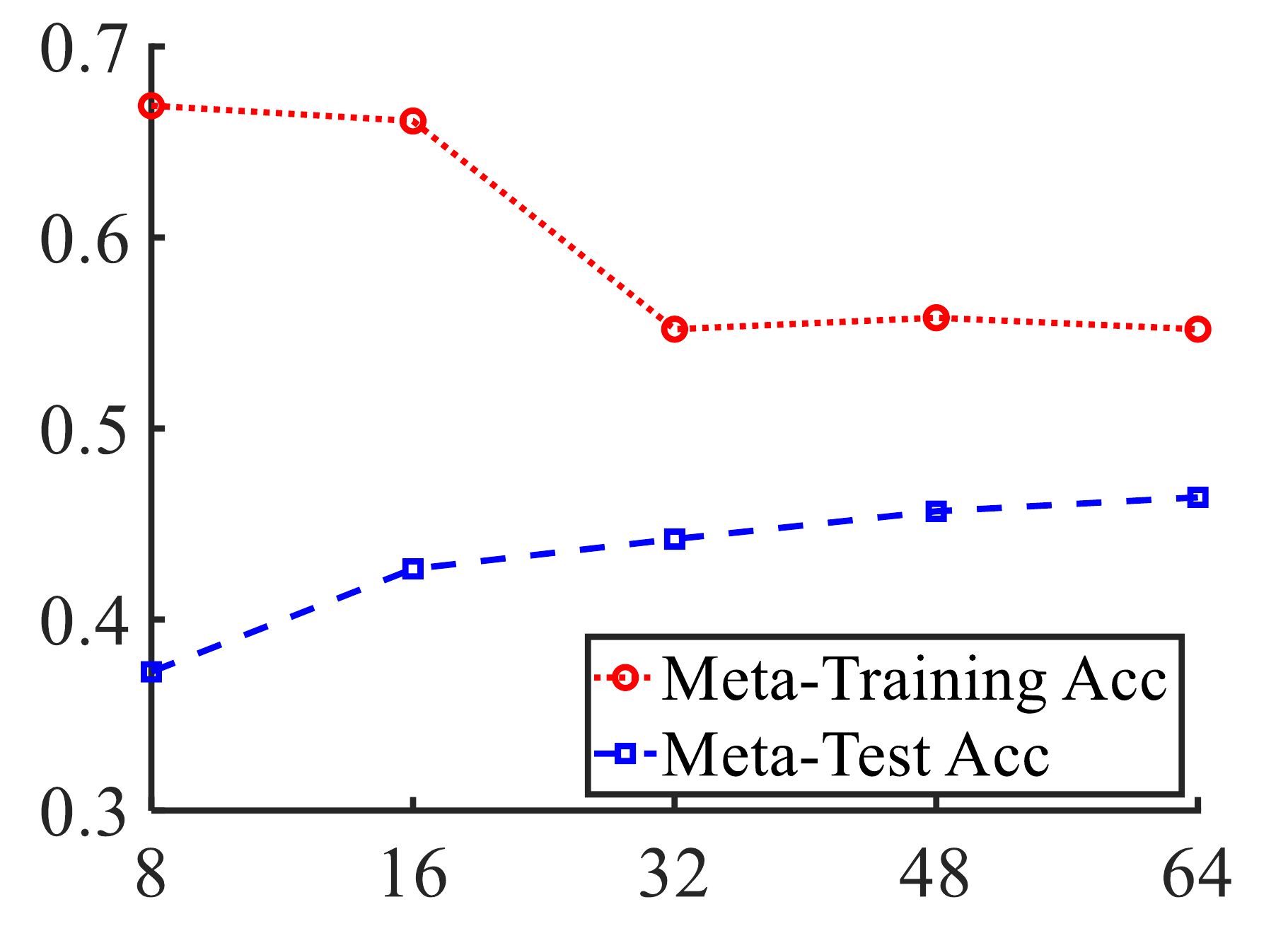}\\
		\mbox{\small ({\it c}) {MatchNet}}
	\end{minipage}
\end{center}
	\caption{1-shot 5-way classification accuracy on {\it Mini}ImageNet (SS). We show meta-training (red) and meta-test (blue) accuracy. We keep all instances per class but vary the numbers of meta-training classes. MAML, ProtoNet, and MatchNet suffer over-fitting when the number of classes gets small. }\label{fig:S_generalization}
\end{figure*}

Besides \textit{Mini}ImageNet and CUB, we further evaluate on the large-scale \textit{tierd}ImageNet data set~\citep{ren2018meta}, which has 779,165 images and 608 classes. We use the standard split: 351/97/160 for meta-train/val/test-pool.

Following the conventional way of image pre-processing~\citep{vinyals2016matching,snell2017prototypical,finn2017model}, we resize all images to $84 \times 84$. 

\paragraph{Meta examples (tasks) and evaluation protocols.} We follow~\citep{rusu2019meta,ye2018learning} to evaluate the few-shot classification by drawing 10,000 tasks from meta-test-pool and there are 15 validation images per class in a task. That is, a meta labeled example $(D_\tr, D_\val)$ has 15 images per class in $D_\val$.
We found the 95\% confidence interval to be consistently within $[0.001, 0.004]$ and thus omit it for brevity.

\paragraph{Baseline methods.} We investigate three popular baselines, namely Model Agnostic Meta-Learning~(MAML)~\citep{finn2017model}, Prototypical Network~(ProtoNet)~\citep{snell2017prototypical}, and Matching Network (MatchNet)~\citep{vinyals2016matching}. MAML implements an inner optimizer to update the meta-learned classifier initializer, and ProtoNet/MatchNet learn discriminative embedding for few-shot classification. ProtoNet uses the distance-based nearest class mean rule for prediction, while MatchNet utilizes the similarity-based nearest neighbor rule.

We re-implement all three algorithms and use the same $C$-way setting in meta-training and meta-test for consistency. That is, we disregard the trick~\citep{snell2017prototypical} that trains with $30$-way tasks but tests with $5$-way tasks.
We apply the first-order MAML and tune the number of updates by meta-validation.  We apply the standard 4-layer ConvNet as the backbone~\citep{vinyals2016matching,snell2017prototypical}. In each layer, convolution, batch normalization, ReLU, and max-pooling are concatenated sequentially. During meta-training SGD with Adam~\citep{KingmaB14ADAM} is employed, with a initial learning rate 2e-3. Our baseline results are close to those reported in the recent overview of few-shot learning~\citep{chen2019closer}.

\begin{figure*}[t]
	\centering
	\footnotesize
	\begin{minipage}[h]{4.2cm}
		\centering
		\includegraphics[width=4.2cm]{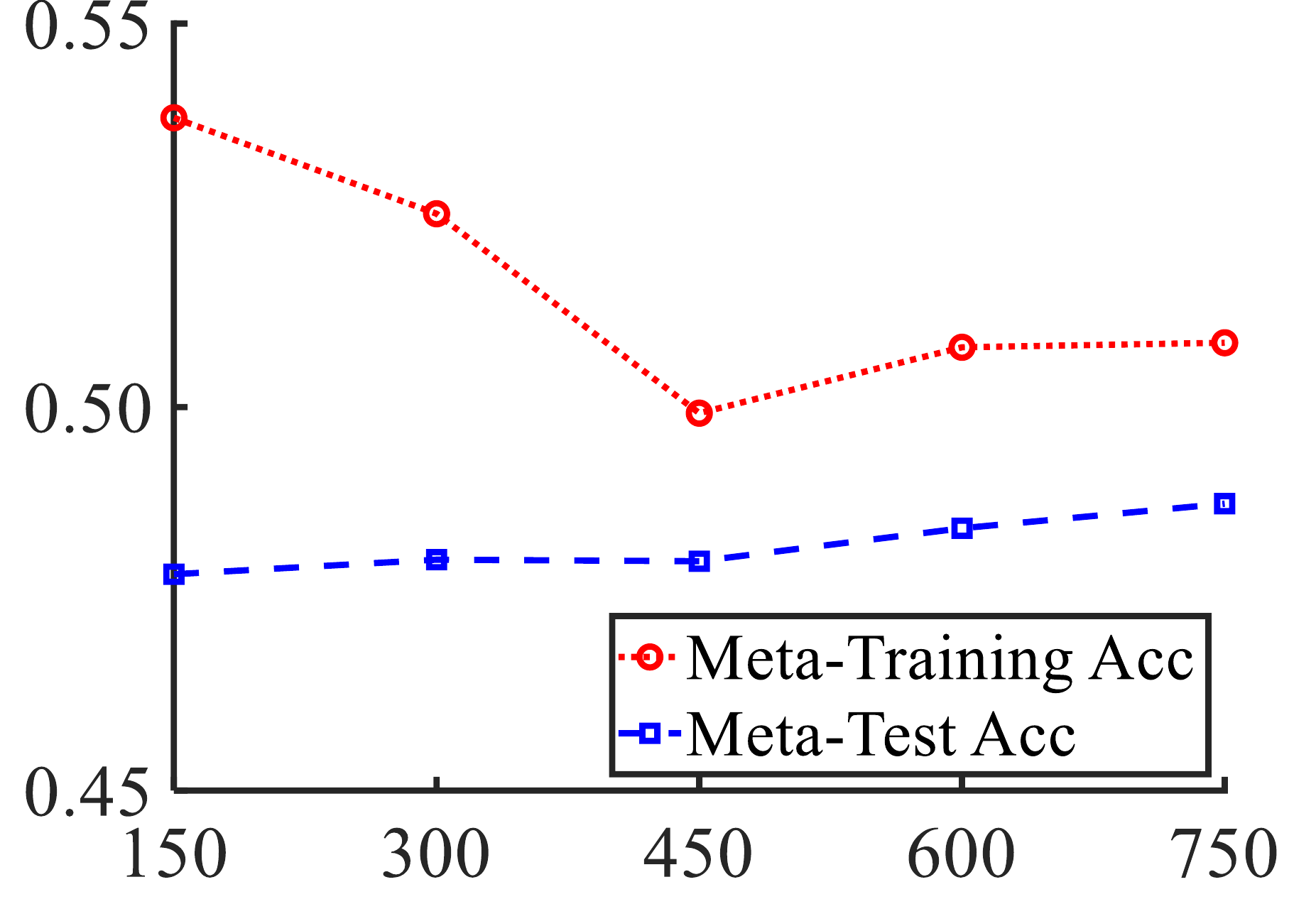}\\
	\end{minipage}
	\hspace{15pt}
	\begin{minipage}[h]{4.2cm}
		\centering
		\includegraphics[width=4.2cm]{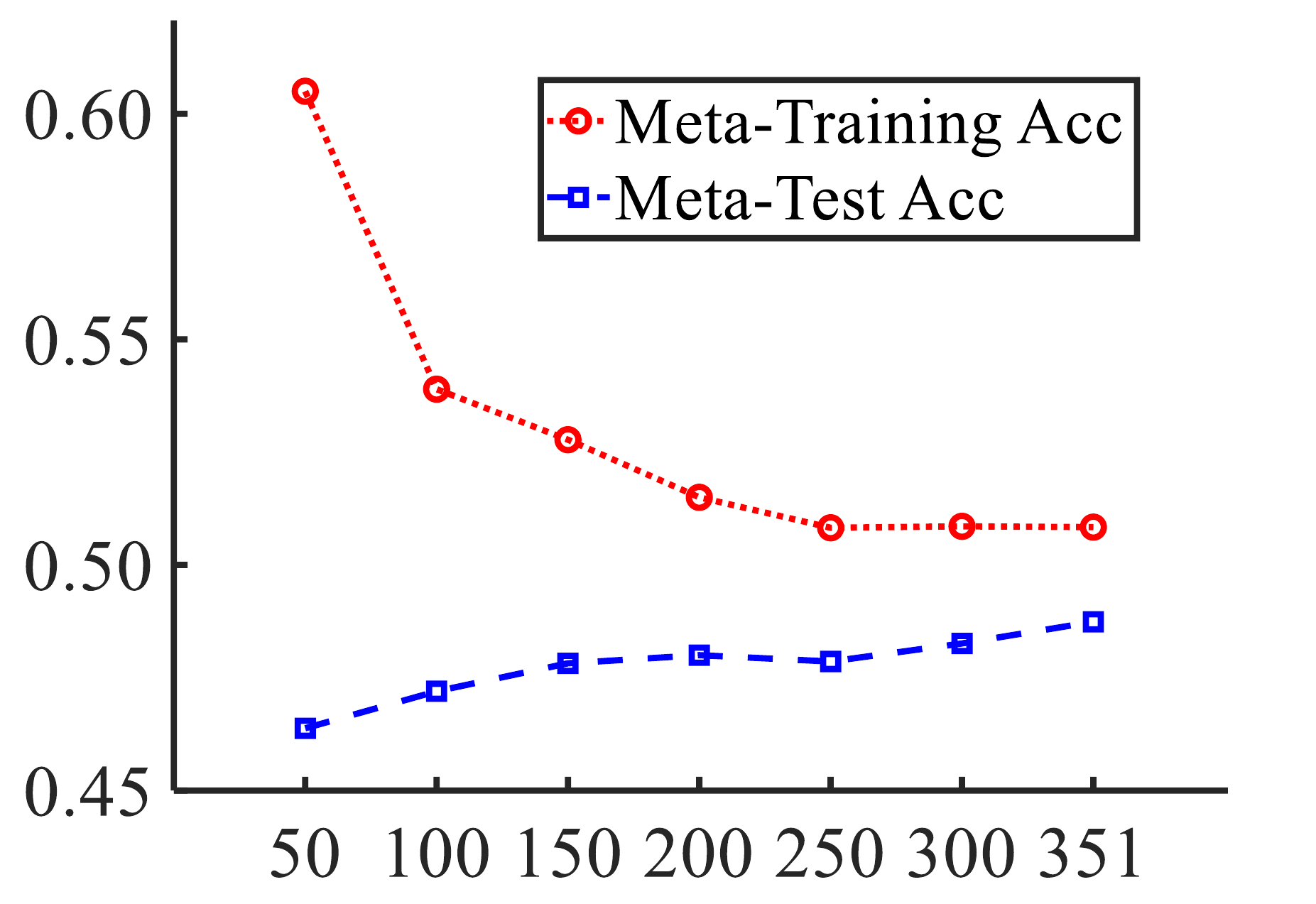}\\
	\end{minipage}
	\caption{1-shot 5-way classification performance of ProtoNet on \textit{tierd}ImageNet: (Left) varying the instances per meta-train class; (Right) varying the number of meta-train classes. By showing meta-training (red) and meta-test (blue) accuracy, we find ProtoNet suffers over-fitting when the number of classes or the number of instances in meta-train classes get small.}
	\label{fig:tierd}
\end{figure*}

\section{Additional Experimental Results}
\label{S_S_exp}
In the following we include more experimental results for the meta-learning analyses.
\subsection{Empirical generalization analysis}
\label{S_Generalization}
Following our supervised view of meta-learning, the generalization ability of the learned meta  model should be affected by the “effective” number of meta training examples. Here we investigate such effective number from another perspective, i.e., the number of classes in the in the meta-train-pool. In detail, we keep all 600 instances per class in {\it Mini}ImageNet (SS), but change the number of available classes for meta-training in the meta-train-pool, from 8 to 64. When there are limited types of classes in the meta-train-pool, the number of diversified tasks to be sampled is constrained. \autoref{fig:S_generalization} shows the change of few-shot accuracy with different meta-train-pools. The trend follows the supervised learning intuition as well: where there are more available meta-training classes (types of few-shot tasks), the better generalization results are achieved. In particular, when there are 8 classes, all models, namely MAML, ProtoNet, and MatchNet overfit. The 1-shot classification performance becomes stable when the number of meta-training classes is larger than 48.

We also conduct the generalization analysis of ProtoNet on the \textit{tierd}ImageNet data set, via varying both the number of instances per meta-train class and the number of meta-train classes. \autoref{fig:tierd} depicts the results, whose trends are similar to \autoref{fig:S_generalization}. 

\begin{table*}[t]
	\centering
	\tabcolsep 2.5pt
	\caption{Bagging for 1/5-shot 5-way classification on {\it Mini}ImageNet (SS). ``w/o Pre-Train" and ``w/ Pre-Train'' denote the vanilla training of the meta-learning model on all classes in the meta-train-pool without or with the pre-training strategy. Single: vanilla training of the meta-learning model on all classes in the meta-train-pool.  ``Bag1'' and ``Bag2'' ensemble the basic models by logits and probabilities respectively. The best result of each shot-algorithm pair is in bold font.}
	\begin{tabular}{c|r|ccc|ccc|ccc}
		\hline
		\multicolumn{2}{c|}{} & \multicolumn{3}{c|}{MAML} & \multicolumn{3}{c|}{ProtoNet} & \multicolumn{3}{c}{MatchNet} \\ \cline{3-11}
		\multicolumn{2}{c|}{} & Single & Bag1 & Bag2 &  Single & Bag1 & Bag2 &  Single & Bag1 & Bag2 \\ \hline
		\multirow{2}{*}{1-shot} & w/o Pre-Train   & 0.459 & 0.448 & 0.482 & 0.461 & 0.488 & 0.491 & 0.463 & 0.481 & 0.482 \\ 
		& w/ Pre-Train    & 0.478 & \textbf{0.498} & \textbf{0.498} & 0.500 & 0.522 & \textbf{0.526} & 0.500 & 0.524 & \textbf{0.525} \\ \hline
		\multirow{2}{*}{5-shot} & w/o Pre-Train  & 0.633 & 0.665 & 0.666 & 0.658 & 0.692 &  0.694 & 0.639 & 0.663 & 0.664\\ 
		& w/ Pre-Train   & 0.660 & \textbf{0.694} & 0.692 & 0.671 & \textbf{0.701} & \textbf{0.701} & 0.642 & \textbf{0.674} & 0.673\\
		\hline
	\end{tabular}
	\label{tab:S_bagging}
\end{table*}

\begin{table*}[htbp]
	\centering
	\caption{The 1-shot and 5-shot classification performance of ProtoNet on the CUB data set. ``w/o Pre-Train" and ``w/ Pre-Train'' denote the vanilla training of the meta-learning model on all classes in the meta-train-pool without or with the pre-training strategy. Average: the averaged few-shot classification results of all basic models.  ``Bag1'' and ``Bag2'' ensemble the basic models with pre-training by logits and probabilities respectively.}
	\begin{tabular}{c|c|cccc@{\;}}
		\addlinespace
		\hline
		& w/o Pre-Train & w/ Pre-Train & Average & Bag1 & Bag2\\
		\hline
		1-Shot & 0.592 & 0.601 & 0.586  & 0.621 & 0.623  \\
		5-Shot & 0.762 & 0.777 & 0.758  & 0.793 & 0.793 \\
		\hline
	\end{tabular}
	\label{tab:S_cub}
\end{table*}

\begin{table*}[t]
	\centering
	\caption{Bagging for 1-shot 5-way classification on {\it Mini}ImageNet (SS) using RN~\citep{sung2018learning} and FEAT~\citep{ye2018learning}. Single: no bagging. Average: average accuracy of basic meta models. Since FEAT already applies pre-training, we do not show its results without pre-training. The best result of each shot-algorithm pair is in bold font.}
    \vskip-7pt
    \begin{tabular}{r|c|cc|c|cc}
	\hline
	 & \multicolumn{3}{c|}{RN} & \multicolumn{3}{c}{FEAT} \\ \cline{2-7}
	 & Single & Average & Bagging &  Single & Average & Bagging \\ \hline
	w/o Pre-Train   & 0.483 & 0.477 & 0.510 & - & - & - \\ 
	w/ Pre-Train    & 0.504 & 0.498 & \textbf{0.529} & 0.552 & 0.545 & \textbf{0.559}  \\
	\hline
\end{tabular}
	\label{tab:bagging_more_meta}
\end{table*}

\begin{table}
	\centering
	\caption{1-shot 5-way classification of ProtoNet on {\it tierd}ImageNet. ``w/o Pre-Train" and ``w/ Pre-Train'' denote the vanilla training of the meta-learning model on all classes in the meta-train-pool without or with the pre-training strategy. ``Bagging'' ensembles the basic models with pre-training by probabilities.} \label{tb:tierd}
	\begin{tabular}{r|c|c}
		\hline
		& Single & Bagging \\ 
		\hline
		w/o Pre-Train &  0.487 &  0.504\\ 
		w/ Pre-Train  &  0.542 & \textbf{0.552} \\
		\hline
	\end{tabular}
\end{table}

\begin{table}[t]
	\centering
	\caption{Bagging for 1-shot 5-way classification on {\it Mini}ImageNet (SS) using ProtoNet with ResNet-12 as the backbone. Single: no bagging. Average: average accuracy of basic meta models.}
\vskip-7pt
\begin{tabular}{r|c|cc}
	\hline
	 & Single & Average & Bagging \\ \hline
	w/ Pre-Train & 0.614 & 0.603 & 0.626  \\
	\hline
\end{tabular}
	\label{tab:bagging_resnet}
\end{table}

\subsection{Supervised learning techniques for meta-learning}
\label{S_Bagging}
In this subsection, we first show different strategies to ensemble the basic meta-learning learner through bagging, and then demonstrate the applicability of bagging to the CUB and \textit{tiered}ImageNet data set. Finally, We experiment with more meta-learning algorithms~\citep{sung2018learning,ye2018learning} and backbones~\citep{he2016deep}. 

We construct bagging models by sub-sampling classes, and build a meta-learning model on the subsets. The differences in various subsets result in the divergence among basic models, which is the key to the success of the ensemble methods. 
To aggregate more than one meta-learning models, we consider averaging their prediction logits (the distance values for ProtoNet and MatchNet) and the normalized probabilities. We denote the two strategies as ``Bag1'' and ``Bag2'', respectively. The results are listed in \autoref{tab:S_bagging}, where bagging show consistent improvements among the vanilla meta-learning models.

To demonstrate its applicability, we also evaluate bagging on another meta-learning benchmark CUB. 
The results of ProtoNet is summarized in \autoref{tab:S_cub}: there exists a consistent trend with the results on the {\it Mini}ImageNet (SS), i.e., models with pre-training work better, and bagging can further improve the performance.

We further evaluate the 1-shot 5-way classification performance of ProtoNet w/ and w/o pre-training and bagging on \textit{tiered}ImageNet. \autoref{tb:tierd} shows the results, which indicates that both techniques are compatible to boost the accuracy.

Finally, we experiment with FEAT~\citep{ye2018learning} and Relational Network (RN)~\citep{sung2018learning} for 1-shot 5-way classification on {\it Mini}ImageNet (SS). We use and adapt their officially released code. We use the standard 4-layer ConvNet as the backbone for a fair comparison. \autoref{tab:bagging_more_meta} shows the results: both pre-training and bagging improve accuracy. We note that we cannot fully reproduce RN's reported accuracy of training from scratch (i.e., $0.504$, vs. ours at $0.483$) without detailed hyper-parameter configurations, but with bagging, we can achieve higher accuracy.

We also experiment with ProtoNet using ResNet-12~\citep{he2016deep,oreshkin2018tadam} as the backbone. We perform pre-training. \autoref{tab:bagging_resnet} shows the results: bagging again leads to improvement. Both results also outperform those reported in~\citep{oreshkin2018tadam} using ResNet-12.

\subsection{Data augmentation}
\label{S_DAug}

\begin{table*}[t]
	\centering
	\caption{Meta-test accuracy of 1-shot 20/30/40-way tasks on {\it Mini}ImageNet (CS). We compare no augmentation (i.e., 1) and augmentation with 8 subcategories by K-means. Since there are only 30 classes in the original meta-train-pool, we cannot perform meta-training with 40-way tasks.}
	\begin{tabular}{c@{\;}|c@{\;}|c@{\;}|c@{\;}|c@{\;}}
		\hline
		\multicolumn{2}{c|}{} & \multicolumn{3}{c}{Meta-Training Tasks}\\ \cline{3-5}
		\multicolumn{2}{c|}{} & 20 & 30 & 40 \\ \cline{3-5}
		\multicolumn{2}{c|}{} & 1 / 8 & 1 / 8 & 1 / 8\\
		\hline
		Meta- & 20 & 0.192  / 0.207 & 0.189 / 0.208 & \hspace{10pt}-\hspace{10pt} / 0.204 \\ 
         Test & 30 & 0.148  / 0.161 & 0.146 / 0.161 & \hspace{10pt}-\hspace{10pt} / 0.159 \\
        Tasks& 40 & 0.122 / 0.134 & 0.120 / 0.134 & \hspace{10pt}-\hspace{10pt} / 0.132 \\
		\hline
	\end{tabular}
	\label{tab:class_aug_1shot}
\end{table*}

\begin{table*}[t]
	\centering
	\caption{Meta-test accuracy of 5-shot 20/30/40-way tasks on {\it Mini}ImageNet (CS). We compare no augmentation (i.e., 1) and augmentation with 8 subcategories by K-means. Since there are only 30 classes in the original meta-train-pool, we cannot perform meta-training with 40-way tasks.}
	\begin{tabular}{c@{\;}|c@{\;}|c@{\;}|c@{\;}|c@{\;}}
		\hline
		\multicolumn{2}{c|}{} & \multicolumn{3}{c}{Meta-Training Tasks}\\ \cline{3-5}
		\multicolumn{2}{c|}{} & 20 & 30 & 40 \\ \cline{3-5}
		\multicolumn{2}{c|}{} & 1 / 8 & 1 / 8 & 1 / 8\\
		\hline
		Meta- & 20 & 0.318 / 0.346 & 0.323 / 0.347 & \hspace{10pt}-\hspace{10pt} / 0.357 \\ 
         Test & 30 & 0.257 / 0.283 & 0.262 / 0.285 & \hspace{10pt}-\hspace{10pt} / 0.294 \\
        Tasks& 40 & 0.220 / 0.244  & 0.224 / 0.245 & \hspace{10pt}-\hspace{10pt} / 0.254 \\
		\hline
	\end{tabular}
	\label{tab:class_aug_5shot}
\end{table*}

We perform K-means within each class to split a class into $K$ subcategories, resulting in $30\times K$ augmented classes in meta-train-pool (and meta-val-pool). To ensure the quality of K-means, we pre-train a 30-way classifier on the whole meta-train-pool and use the features for K-means. We run 30 trials of K-means to compensate its randomness. We construct a meta-training example (task) by first picking a trial and sampling 30 classes from the corresponding augmented meta-train-pool.

Here we provide additional results on {\it Mini}ImageNet (CS).
We further evaluate on 20-way and 40-way tasks during meta-test and also report 5-shot classification results. We note that since ProtoNet~\citep{snell2017prototypical} learns a feature extractor $\psi$ (cf. \autoref{e_ProtoNet} of the main text) during meta-training, it can be applied to a task of different ways (e.g., meta-training with 20-way tasks but meta-test with 40-way tasks).

\autoref{tab:class_aug_1shot} and \autoref{tab:class_aug_5shot} summarize the results. We consider no augmentation or augmentation with 8 subcategories by K-means and report different combinations of meta-training and meta-test tasks. We can clearly see the trend (within each column): data augmentation consistently improves the performance by generating more diverse meta-training examples. 

Moreover, by comparing within each row where the meta-training tasks are different, we see no clear performance drop when the meta-training and meta-test tasks are different. Interestingly, for 5-shot tasks (cf. \autoref{tab:class_aug_5shot}) meta-training with more ways seem to be always beneficial, which aligns with the observation in~\citep{snell2017prototypical}. A more detailed investigation is thus desirable in the future work. 

\subsection{Non-parametric methods}
We introduce meta-KNN as a widely-applicable framework to apply non-parametric methods to meta-learning. Here we provide the hyper-parameters we use. We set $K=100$, $\beta=1$, and $\alpha=0.0002$  (cf. \autoref{a_meta_KNN} of the main text).
$\alpha=0.0002$ is the learning rate when the initial ProtoNet $\hat{g}$ converges. 

For a test task, meta-KNN takes 5 ms for searching KNN tasks and 360 ms for fine-tuning (using one RTX 2080 GPU), which is quite efficient.
We note that meta-KNN (\autoref{a_meta_KNN}) can easily incorporate various kinds of metrics for KNN search, and we applied the simple average features with Euclidean distance as an illustrative example. We expect that applying a more sophisticated and application-dependent metric (e.g., using higher-order statistics~\citep{Achille2019Information} or Task2Vec~\citep{Achille2019Task}) to characterize task similarity would further improve the performance.

\begin{table*}[t]
	\centering
	\tabcolsep 2pt
	\caption{Few-shot 5-way domain generalization on Office-Home with two domains: Clipart (C), Product (P).
	}
	\begin{tabular}{r|cc|cc|cc}
		\hline
		& \multicolumn{2}{c|}{Meta-train} & \multicolumn{2}{c|}{Meta-test} & \multicolumn{2}{c}{Test Acc}\\ \cline{2-7}
		Case & Source & Target & Source & Target & 1-Shot & 5-Shot \\ \hline
		I-1 & C     & C     & C     & C     & 0.341 & 0.477 \\ 
		I-2 & P     & C     & P     & C     & 0.296 & 0.350\\
		I-3 & C     & C     & P     & C     & 0.275 & 0.342\\
		I-4 & P     & P     & P     & C     & 0.264 & 0.283\\
		\hline
	\end{tabular}
	\hfill
	\begin{tabular}{r|cc|cc|cc}
		\hline
		& \multicolumn{2}{c|}{Meta-train} & \multicolumn{2}{c|}{Meta-test} & \multicolumn{2}{c}{Test Acc}\\ \cline{2-7}
		Case & Source & Target & Source & Target & 1-Shot & 5-Shot \\ \hline
		II-1 & P     & P     & P     & P     & 0.448 & 0.609\\ 
		II-2 & C     & P     & C     & P     & 0.291 & 0.381\\
		II-3 & C     & C     & C     & P     & 0.284 & 0.338\\
		II-4 & P     & P     & C     & P     & 0.275 & 0.329\\
		\hline
	\end{tabular}
	\label{tab:do2}
\end{table*}

\subsection{Domain generalization and meta Domain shifts}

This section shows more experimental results for domain generalization and meta domain shifts.
We experiment on the Off-Home dataset, which is designed for domain adaptation. We focus on two domains: Clipart (C) and Product (P). There are 65 classes and we split them into 25 for meta-train-pool, 15 for meta-val-pool, and 25 for meta-test-pool. We work on \emph{few-shot 5-way} tasks in which a task contains a training set $D_\tr$ from one (source) domain and a validation set $D_\val$ from the other (target) domain.

\autoref{tab:do2} summarizes the results. We clearly see the difficulty when the source and target domains within a task are different by comparing Case I-1 and I-2 (similarly, II-1 and II-2). We note that in each of these cases the distribution of the meta-training and meta-test examples are the same. Therefore, there is no meta-domain shift and the gap simply indicates the difficulty of tasks.

We further investigate meta-domain shifts by constructing meta-training and meta-test examples in a different way (Case I-3, I-4 and II-3, II-4). We note that some meta-learning algorithms, by default, cannot learn to handle the case where $D_\tr$ and $D_\val$ are sampled from different domains (e.g.,~\citep{nichol2018first}). These additional experiments are meant to simulate the results by those algorithms.
The performance in these cases is outperformed by Case I-2 and II-2 respectively where each group considers the same meta-test examples. We argue that this gap indeed results from meta-domain shifts (cf. \autoref{S_general} in the main paper) and might be resolved via domain adaptation by casting meta-learning as supervised learning.

\begin{table}
	\centering
	\tabcolsep 1pt
	\caption{20-shot 5-way domain generalization on Office-Home with two domains: Clipart (C) and Product (P) domains.} 	\label{tab:S_domain}
	\begin{tabular}{c|cc|cc|c}
		\hline
		& \multicolumn{2}{c|}{Meta-train} & \multicolumn{2}{c|}{Meta-test} & \multicolumn{1}{c}{Test Acc}\\ 
		\cline{2-6}
		Case & Source & Target & Source & Target & 20-Shot \\ 
		\hline
		I-1 & C     & C     & C     & C     & 0.554  \\ 
		I-2 & P     & C     & P     & C     & 0.358 \\
		I-3 & C     & C     & P     & C     & 0.349 \\
		I-4 & P     & P     & P     & C     & 0.312 \\
		\hline
	\end{tabular}
\end{table}

\begin{table}
	\centering
	\caption{20-shot 5-way domain generalization (the I-2 case) with pre-training and bagging.} \label{tb:DG}
	\begin{tabular}{r|c|c}
		\hline
		& Single & Bagging \\ 
		\hline
		w/o Pre-Train &  0.358 &  0.383\\ 
		w/ Pre-Train  &  0.377 & \textbf{0.392} \\
		\hline
	\end{tabular}
\end{table}

We further extend our studies of domain generalization beyond the few-shot learning setting.
For a domain generalization task, a base model could be trained on labeled instances from one (source) domain, not necessarily few-shot, and tested on instances from another (target) domain. Thus, besides the 1-shot and 5-shot settings as shown before, we examine a many-shot setting (e.g., 20-shot) with ProtoNet on a domain generalization task.

\autoref{tab:S_domain} shows the results: the gap between I-2 and I-3 (I-4) reflects the meta-domain difference between meta-train and meta-test tasks (irrespective of 1- or 20-shot), similar to the problem of domain adaptation in supervised learning. For I-2 (the source and target domains are Product and Clipart respectively in both meta-train and meta-test) we further apply pre-training and bagging. As shown in \autoref{tb:DG}, both techniques are applicable to the domain generalization tasks.
\end{document}